%% file: main.tex
\documentclass[10pt]{article} % For LaTeX2e
\usepackage[accepted]{tmlr}
\input{math_commands.tex}

\usepackage{hyperref}
\usepackage{url}

\title{A Greedy Hierarchical Approach to Whole-Network Filter-Pruning in CNNs}

\author{\name Kiran Purohit \email kiran.purohit@kgpian.iitkgp.ac.in \\
\name Anurag Reddy Parvathgari \email anuragparvathgari7@gmail.com \\
\name Sourangshu Bhattacharya \email sourangshu@cse.iitkgp.ac.in\\
\addr Department of Computer Science and Engineering\\
Indian Institute of Technology, Kharagpur, India}

\def\month{08}  % Insert correct month for camera-ready version
\def\year{2024} % Insert correct year for camera-ready version
\def\openreview{\url{https://openreview.net/forum?id=WzHuebRSgQ}} % Insert correct link to OpenReview for camera-ready version

\newcommand{\search}{\texttt{HBGS}}
\newcommand{\searchb}{\texttt{HBGS-B}}
\newcommand{\tree}{\texttt{HBGTS}}
\newcommand{\treeb}{\texttt{HBGTS-B}}

\newcommand{\fpback}{\texttt{FP-Backward}}

\usepackage{multirow}
\usepackage{hhline}
\usepackage{color}
\usepackage{wrapfig}

\usepackage{mathtools}
\usepackage{microtype}
\usepackage{graphicx}
\usepackage{subfigure}
\usepackage{booktabs} % for professional tables
\usepackage{bigstrut}
\usepackage{hyperref}

\usepackage{mathtools}
\usepackage{amsthm}
\usepackage{bbm}
\usepackage[capitalize,noabbrev]{cleveref}
\usepackage{algpseudocode}
\usepackage{algorithm}
\usepackage{xcolor}
\usepackage{color}
\usepackage{natbib}

\usepackage{algpseudocode}
\usepackage{algorithm}

\newtheorem{theorem}{Result}
\newcommand{\RR}{\mathbb{R}}
\newcommand{\cD}{\mathcal{D}}

\begin{document}

\maketitle

\begin{abstract}
Deep convolutional neural networks (CNNs) have achieved impressive performance in many computer vision tasks. However, their large model sizes require heavy computational resources, making pruning redundant filters from existing pre-trained CNNs an essential task in developing efficient models for resource-constrained devices.
Whole-network filter pruning algorithms prune varying fractions of filters from each layer, hence providing greater flexibility.
State-of-the-art whole-network pruning methods are either computationally expensive due to the need to calculate the loss for each pruned filter using a training dataset, or use various heuristic / learned criteria for determining the pruning fractions for each layer.
Hence there is a need for a simple and efficient technique for whole network pruning.
This paper proposes a two-level hierarchical approach for whole-network filter pruning which is efficient and uses the classification loss as the final criterion. 
The lower-level algorithm (called filter-pruning) uses a sparse-approximation formulation based on linear approximation of filter weights. 
We explore two algorithms: orthogonal matching pursuit-based greedy selection and a greedy backward pruning approach.
The backward pruning algorithm uses a novel closed-form error criterion for efficiently selecting the optimal filter at each stage, thus making the whole algorithm much faster.
The higher-level algorithm (called layer-selection) greedily selects the best-pruned layer (pruning using the filter-selection algorithm) using a global pruning criterion. We propose algorithms for two different global-pruning criteria: (1) layerwise-relative error (HBGS), and (2) final classification error (HBGTS).
% uses an activation reconstruction-based criteria along with the output from the lower-level algorithm to identify the layer to prune filters from.
% We further improve the efficiency of the lower-level algorithm by proposing a backward elimination algorithm using novel closed-form error criteria. 
Our suite of algorithms outperforms state-of-the-art pruning methods on ResNet18, ResNet32, ResNet56, VGG16, and ResNext101. 
Our method reduces the RAM requirement for ResNext101 from 7.6 GB to 1.5 GB and achieves a 94\% reduction in FLOPS without losing accuracy on CIFAR-10.
Our code is available \textbf{here} \footnote{https://github.com/filter-pruning/Hierarchical-Filter-Pruning}
\footnote{Sourangshu Bhattacharya is thankful to ANRF Core Research Grant No. CRG/2023/004600}
\end{abstract}

\input{intro}
\input{related}
\input{method}

\input{experiment}

\input{conclude}

\bibliography{ref}
\bibliographystyle{tmlr}

\appendix
\newpage
\input{proofs}
\input{supple}
\end{document}

%% file: math_commands.tex
\usepackage{amsmath,amsfonts,bm}

\def\eqref#1{equation~\ref{#1}}
\def\1{\bm{1}}

\DeclareMathAlphabet{\mathsfit}{\encodingdefault}{\sfdefault}{m}{sl}
\SetMathAlphabet{\mathsfit}{bold}{\encodingdefault}{\sfdefault}{bx}{n}

%% file: intro.tex
\section{Introduction}
\label{sec:intro}
% \textbf{-----SB: Introduction and abstract is remaining-----}

% \color{blue}
%  Lottery ticket requires training from scratch and does not retain the information present in pre-trained models. Hence these approaches are extremely resource-intensive at training time, both in terms of labeled data, and computational requirements. On the other hand, filter pruning techniques start with an existing pre-trained model, and try to prune away redundant filters in each layer. However, these techniques suffer from two drawbacks:\\
% (1) Many times, the criteria used for pruning filters are based on the filter weights, rather than the output of various layers on a given input dataset.\\
% (2) These techniques prune uniformly / proportionately from each layer. However, certain layers may have a higher proportion of redundant filters.

% In this work, we design a novel hierarchical search-based sparse approximation scheme for filter pruning.\\

% Our Contributions:\\
% - We pose the problem of Linearly Replaceable Filters as a vector-sparse approximation problem.\\
% - We pose the problem of designing an algorithm for (layer, filter) - selection as a hierarchical greedy search problem.\\
% - To speed up the filter selection of the hierarchical search method, we propose a backward elimination-based algorithm.\\

% Highlights of Experimental Results:\\
% -On CIFAR100 at 98\% parameter reduction we are having 4\% and 3.8\% improvement over LRF on ResNet56 and VGG16 respectively. \\
% -On CIFAR10 at 95\% parameter reduction we are having 3.2\% improvement over LRF on ResNet18.

Convolutional neural networks (CNNs) have demonstrated remarkable performance across various applications, such as image classification \citep{han2015deep}, object detection \citep{redmon2016you}, and image segmentation \citep{minaee2021image}. However, the deployment of CNNs on IoT devices for computer vision tasks often encounters practical bottlenecks related to the model size and computational complexity of inference (FLOPs) \citep{goel2020survey}. 
While neural architecture search \citep{baker2016designing,zoph2016neural} and efficient model design \citep{tan2019efficientnet} can sometimes lead to highly efficient architectures, they impose substantial requirements in terms of data and computational cost, as well as research expertise.  However, pruning of pre-trained models \citep{lebedev2018speeding,  hoefler2021sparsity, vadera2022methods,  he2023structured}  provides a cheaper alternative where one can avoid re-training complicated models on large datasets. 
For CNNs, \textit{structured pruning} or \textit{filter-pruning} (FP) \citep{he2017channel, luo2017thinet, he2023structured} has emerged as a preferred alternative since it causes a reduction in computation (thus leading to power savings) as well as memory requirements without requiring special hardware or re-implementation of operations.

Filter-pruning (FP) techniques can further be classified as (1) \textit{layer-wise pruning}, which prune filters uniformly from each layer (e.g. score-propagation \citep{yu2018nisp} and error in activation reconstruction \citep{luo2017thinet}), and (2) \textit{whole-network pruning} (WNP), which prunes filters from the entire network. The WNP approach can prune different fractions of filters from each layer, hence providing higher flexibility. 
% WNP is preferred more compared to layer-wise pruning as the final prediction is the result of collaborative contributions from all layers in the network.
%However, this can be computationally expensive for large DNNs.
An important challenge for WNP is to determine the pruning fractions for each layer.
\citep{kuzmin2019taxonomy} (section 3.4.1) calculates the accuracy by pruning individual layers to a varying fraction and finds an optimal compromise so that the overall pruning ratio is achieved while minimizing the maximum loss in accuracy per layer. The main disadvantage of this approach is that the effect of pruning one layer on the pruning of another layer is not captured. 
% Recent WNP techniques have two major limitations. They are either expensive/inefficient or use some approximate criteria to prune filters. 
Recent works also include methods based on Taylor-series expansion of the loss function, which approximates the influence score of pruning a filter on the overall loss function \citep{wang2019eigendamage, molchanov2019importance, peng2019collaborative, nonnenmacher2021sosp} with good practical performance \citep{he2023structured}. 
However, these methods can be expensive for large networks as they require passing over the entire training set to calculate each influence score, which can be costly for large datasets. 
Additionally, \citep{dong2019network} applied NAS to search for a network with flexible channel and layer sizes, but this method can also be expensive for larger networks. 
On the other hand, some recent works use approximate criteria to prune filters. For instance, \citep{murti2022tvsprune} propose a discriminative pruning technique based on total variation separation distance (TVS), which is an approximate criterion to prune filters from a network. Similarly, \citep{he2020learning} choose different criteria to prune filters from different layers using Gumbel-softmax. However, the main drawback of this procedure is that the Gumbel-softmax smoothing only calculates an approximate output feature map for each layer, thus potentially hurting overall performance. 
Therefore, there is a need for an \textit{efficient} and \textit{accurate} WNP technique that directly optimizes the \textit{training data loss}.
% \color{red}, need to write why training loss is important. \color{black}

% \citep{dong2019network} learns the probability of sampling each filter in each layer by constructing a weight-aggregated feature output for each layer from the sampled feature maps. 
% The overall algorithm is end-to-end differentiable using Gumbel-softmax smoothing. They learn the probability parameters by minimizing a combination of validation loss and computational cost. The main drawback of this procedure is that the Gumbel-softmax smoothing only calculates an approximate output feature map for each layer, thus hurting overall performance. 

% Existing methods use various criteria:\\
%such as  NN-based scoring \citep{he2020learning,dong2019network},  PQ-index \citep{diao2022pruning}, etc. 
%\color{red} However, these heuristic criteria do not necessarily lead to optimal performance.
% \color{blue}
% Another class of methods globally optimizes specific criteria over all the filters of a network: e.g. 
% error in activation reconstruction \citep{chen2022otov2, el2022data}, 
% \citep{sui2021chip}  Channel independence-based pruning UNIFORM PRUNING \\
% \citep{diao2022pruning} UNSTRUCTURED \\

% \color{red} Since naive optimization methods are not scalable to many modern networks, a key open problem is to design a scalable algorithm.
% \citep{chen2022otov2} uses a projected gradient descent algorithm, potentially leading to local optima, 
%while \citep{el2022data} uses a greedy submodular optimization-based method for selecting filters in each layer. 

\color{black}

\begin{figure}[t]
\centering
\includegraphics[width=0.99\textwidth]{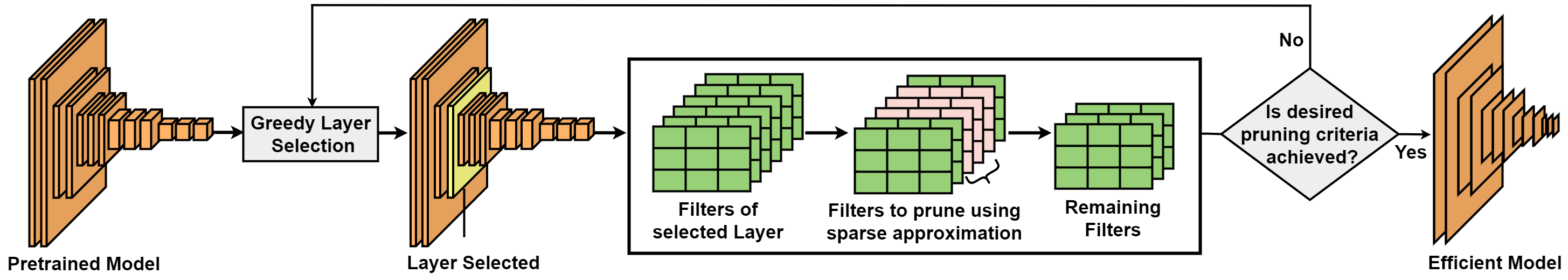}\\
\caption{Hierarchical approach for non-uniform pruning of filters across the network. 
% A pre-trained model is passed for pruning to our \search/\tree\ method. The first block of this method selects the prospective filters to prune using sparse approximation from each layer across the network. Then the second block makes use of the prospective remaining weights of each layer to calculate their reconstruction error. A layer with the minimum error is selected to prune filters from. Then we run one epoch of fine-tuning after the pruning of the filters from the selected layer. The fine-tuned model is sent again to the first block and this process continues till we reach the desired sparsity. Finally, we get an accurate, efficient, and compressed model.
}
\label{fig:block_diagram}
% \vspace{-4mm}
\end{figure}
In this work, we propose a greedy hierarchical training data loss-based approach for whole-network filter pruning (see fig. \ref{fig:block_diagram}).
The iterative higher-level algorithm (called \textit{layer-selection}) evaluates all layers based on outputs from the lower-level algorithm, and greedily selects a layer to prune filters from in each iteration.
The lower-level algorithm (called \textit{filter-pruning}) prunes filters optimally for the current network configuration.
We propose two versions of the iterative layer-selection algorithm: (1) \textbf{hierarchical backward greedy search (\search)}, which selects layers based on the relative reconstruction error of the layer outputs, and (2) \textbf{hierarchical backward greedy tree search (\tree)} which selects the layers based on the error of the final classification layer outputs.
The key advantage of our greedy layer-selection, compared to a learned criterion \citep{he2020learning}, or a threshold-based criterion \citep{kuzmin2019taxonomy} is that we utilize the activations from the modified network, which arguably leads to better decisions after a few iterations of pruning. 
However, since each iteration of the greedy layer-selection makes many calls to the filter-pruning algorithm (typically, the number of layers many calls, with some possibility of caching old results), an expensive filter-pruning algorithm would be impractical for large networks.

A key contribution of this paper is to propose an \textit{efficient} filter-pruning algorithms, which can ensure the feasibility of the overall hierarchical scheme for large networks.
% Layer-wise filter-selection techniques broadly use two types of criteria: (1) \textit{weight-based} e.g. pruning linearly replaceble filters (LRF) \citep{joo2021linearly}, median weight-norm based \citep{he2019filter}, etc, and (2) \textit{layer output-based} \citep{luo2017thinet,sui2021chip,el2022data} which minimize the error in output from the layers. Layer-output based methods are computationally expensive, since they need to compute the outputs using a training dataset.
While LRF \citep{joo2021linearly} demonstrated impressive practical performance for pruning filters, it only prunes one filter at-a-time, hence making it prohibitively expensive for our purpose.
We formulate the problem of optimally pruning multiple filters from a layer using linear replaceability criteria as a sparse approximation problem.
We study an orthogonal matching pursuit (OMP) \citep{tropp2007signal} based algorithm, FP-OMP \citep{10.1145/3564121.3564139} for filter-pruning. Under the assumption of restricted isometry of the matrix composed of filter weights \citep{tropp2007signal}, FP-OMP selects filters whose linear combinations can represent the pruned filters with minimal error.
However, since FP-OMP follows a greedy forward selection algorithm, and is called iteratively 
to prune a small number of filters, the overall algorithm becomes computationally expensive.
To alleviate this inefficiency, we propose \textbf{\fpback} -- a backward elimination-based algorithm for solving the sparse approximation problem. A key facilitating factor towards a fast implementation of \fpback\ is the calculation of a closed-form expression for the approximation error incurred by pruning a filter. \tree\ along with \fpback\ (called \textbf{\treeb}) is an \textit{efficient} algorithm taking only 50\% of the running time of \tree\ and with performance comparable to it. 

\color{black}

Experimental results on a variety of standard pre-trained CNN models, e.g. ResNet18, ResNet32, ResNet56, VGG16, ResNext101 on standard image classification datasets, e.g. CIFAR10, CIFAR100, Tiny-Imagenet show that models pruned with \search\ and \tree\, have higher accuracies compared to recent state-of-the-art pruning methods for similar compression levels (see Table \ref{tab:VGG16_ResNet18} and Figure \ref{fig:line_graph}). At higher parameter reduction ($\geq$90\%), the proposed methods outperform existing methods by $\sim 5\%$ (see Figure \ref{fig:line_graph}). We also find the optimal pruned model to have a highly non-uniform pruning fraction distribution for each layer (see Figure \ref{fig:prune_percent_errror_reconst}), hence showing the effectiveness of our layer-selection algorithm. To summarize:
\begin{enumerate}
    \item We propose a novel \textit{greedy hierarchical framework} for non-uniform pruning of filters with \textit{filter-pruning} at the lower level and \textit{layer-selection} at a higher level. 
    \item We propose a backward-elimination-based scheme, \fpback\ for filter pruning, which takes advantage of a novel closed-form expression for approximation error.
    \item We propose \tree\, which uses an efficient implementation to directly optimize the classification error for layer selection.
\end{enumerate}

\color{black}

%% file: related.tex
\section{Related Work}
\label{sec;related}

Many pruning methods have been proposed in the literature. \citep{  hoefler2021sparsity, vadera2022methods} provide excellent surveys for pruning techniques. 
% each using different selection schemes to solve complexity issues \citep{lebedev2018speeding,  hoefler2021sparsity, vadera2022methods,  he2023structured}.
Pruning can be categorised two types: \textit{unstructured pruning}, involving the removal of individual weights \citep{han2015learning}, and \textit{structured pruning} or \textit{filter-pruning} (FP), in which entire nodes or channels are removed \citep{he2017channel, luo2017thinet, he2023structured}. Structured pruning provides efficiently implementable models on a wide range of accelerator devices e.g. GPUs.
 \citep{he2023structured} provides a recent survey and website for comparing structured pruning techniques for CNNs.
% A pre-trained neural network can maintain its performance even after removing unnecessary weights \citep{lebedev2018speeding,  hoefler2021sparsity, vadera2022methods,  he2023structured}. 
Pruning can be done on a pre-trained model or from scratch, which is costly and requires large training data. Therefore we focus on pruning a pre-trained model. We further categorise it into the following groups: \\
\textbf{Weight-Based Pruning -} Weights of filters are examined to determine which ones are essential for the model's performance. These methods do not require input data.
\citep{han2015learning} focused on eliminating small-norm weights. \citep{he2019filter} incorporates geometric median \citep{fletcher2008robust} to estimate the importance of each filter. \citep{joo2021linearly}  prunes linearly replaceable filters. \\
\textbf{Activation-Based Pruning -} Rather than relying on filter weights, these methods utilize activation maps or layer outputs to make pruning decisions. We can utilize information from activation maps at the \textit{current layer} or \textit{all layers/whole-network}. Some of the \textit{current layer} activation-based pruning methods are: CP \citep{he2017channel} which focuses on minimizing the reconstruction error of sparse activation maps, while HRank \citep{lin2020hrank} calculates the average rank of activation maps. CHIP \citep{sui2021chip} assesses cross-channel correlation to evaluate channel importance. ThiNet \citep{luo2017thinet, el2022data} approximates activation maps of layer l+1 using subsets of layer l's activation maps. \textit{Current layer} activation-based pruning methods do not consider the reconstruction error propagation. Some of the \textit{all layers/whole-network} activation-based pruning methods are: NISP \citep{yu2018nisp} which assesses the Final Response Layer to calculate the neuron importance, while DCP \citep{zhuang2018discrimination} aims to retain discriminative channels. Layer-output based methods are computationally expensive, since they need to compute the outputs using a training dataset. \\
\textbf{Regularization -}
Regularization can aid in learning structured sparse networks by incorporating various sparsity regularizers. These regularizers can be implemented on \textit{Batch Normalization (BN) parameters} \citep{liu2017learning, you2019gate, kang2020operation}, with \textit{extra parameters} \citep{huang2018data,lin2019towards} and \textit{filters} \citep{wen2016learning,chen2021only}. \\
\textbf{Taylor Expansion -} Taylor Expansion is employed to approximate the change in the loss by pruning filters.
\textit{First-order-Taylor} uses the first-order information to calculate the loss change caused by pruning weights \citep{molchanov2019importance, you2019gate}. \textit{Second-order-Taylor} exploits the Hessian matrix, containing second-order information \citep{peng2019collaborative,wang2019eigendamage, liu2021group}. However, these methods can be expensive for large networks.
\\
\textbf{Coreset-Based Pruning -} Recently, several methods \citep{tukan2022pruning,liebenweinprovable,mussay2021data} have utilized the concept of coresets for pruning DNNs. These approaches focus on providing bounds on the approximation error incurred at each layer of prediction. However, they do not necessarily achieve optimal pruning in terms of representation error. \cite{tukan2022pruning} employs an upper bound on sensitivity for sampling filters, which can result in overly pessimistic sampling weights. On the other hand, our method focuses on removing linearly redundant filters, which is done optimally for a given layer.

There is a different line of work in which pruned models are effectively trained from \textit{scratch} e.g. \cite{frankle2018lottery,rachwan2022winning}. Unlike training-based approaches, our method does not require training from scratch, which is costly and requires large training data.

%% file: method.tex
\color{black}

% \section{Hierarchical Sparse Approximation and Backward Greedy Search for Filter Pruning}
\section{A Hierarchical Greedy Approach to Filter Pruning}
\label{sec:methods}

We propose a Hierarchical scheme, \search/\tree\, for non-uniform filter pruning from a pre-trained CNN. As shown in Figure \ref{fig:block_diagram}, the proposed scheme operates in a two-level hierarchical manner: (1) \textit{filter pruning} - at the lower level, this step identifies the most appropriate filters to be pruned from each layer and (2) \textit{layer selection} - at a higher level this step selects the best layer to currently prune from. These two steps are applied iteratively to achieve a non-uniform pruning from the whole network. 
%the setup for filter pruning/selection for the architecture of Convolutional Neural Networks. We lay the background of the problem in 
We first describe our sparse approximation-based formulation for optimal filter pruning from each layer, and then describe a faster backward elimination-based algorithm for the same.
For layer selection, we describe a layerwise-regression-based backward greedy search strategy. We also incorporate an overall error-based strategy for layer selection.
% where we describe the limitations of the existing State-of-the-art methodology. This serves as a motivation for our proposed pruning framework in Section \ref{sec:background}. Inspired from \citep{joo2021linearly}, we derive the weight compensation module for our setup in Section \ref{sec:weightcomp}. 

\color{black}

\subsection{Sparse Approximation for Filter Pruning}
\label{sec:background}

A convolutional filter used in deep CNN is denoted by a $K \times K$ matrix. A convolutional layer is defined as filter weights $ f_{i,j} \in \RR^{K^2}$, where $i=1,...,m$ and $j=1,...,n$ are the number of input and output channels.
Given the input feature map with $m$ channels $X = \{X_1 , ... , X_m \}$, the output feature map with $n$-channels $Y = \{Y_1 ,..., Y_n \}$, can be calculated as: $Y_j = \sum_{i=1}^m X_i \ast f_{i,j} := X \ast f_{:,j}$ 

Here, $\ast$ denotes the convolution operation, and $f_{:,j} \in \RR^{K^2\times m}$ denotes all the filter weights for output channel $j$. For brevity, we describe the algorithm for output channel pruning throughout the paper. Input channel pruning can be performed analogously.  For channel pruning, we follow the idea of \textit{linearly replaceable filters} (LRF) introduced in \citep{joo2021linearly}, which states that any filter $f_{:,j} \in \RR^{K^2m}$ can be pruned if the filter weights can be expressed as a linear combination of other filter weights of the same layer which are not pruned. Note that, for linear approximation of a filter with respect to other filters of the same layer, we treat the filter weights, $f_{:,j}$, as a flat $K^2m$-dimensional vector. For LRF, we prune the channel $j$ such that $\|\epsilon_j\|$ is minimum, where $f_{:,j} = \sum_{l \neq j} \lambda_{j,l}f_{:,l} + \epsilon_j $. Here, $\lambda_{j,l}$ are the coefficients of $l^{th}$ filter for approximating the $j^{th}$ filter, which can be computed by solving the minimization problem: $\min_{\lambda_{j,:}} ||f_{:,j} - \sum_{l \neq j} \lambda_{j,l} f_{:,l}||^2 $. LRF \citep{joo2021linearly} works by iteratively pruning one filter using the above technique, and updating the weights of the retained parameters using one epoch of SGD updates minimizing a combination of training loss and knowledge distillation loss \citep{hinton2015distilling} w.r.t. to unpruned model outputs.

\begin{figure}[t]
\begin{minipage}{0.46\textwidth}
\begin{algorithm}[H]
 \caption{Filter Pruning-OMP (FP-OMP)}
 \label{algo:omp_sparse}
 % \footnotesize
 \begin{algorithmic}[1]
 
 \State \textbf{Input:} $n$: Number of filters, \State \hspace{11mm} $\beta$: Pruning fraction, 
 \State \hspace{11mm} $f_{:,j} \in \RR^{K^2m}$ \hspace{-1mm} $j=1,....n$:  Filters 
 
 \State \textbf{Initialize:}
 \State \hspace{2mm}  Normalize $f_{:,j}$ such that $||f_{:,j}||_2 = 1$ 
\State \hspace{2mm} $R_j = f_{:,j} \hspace{2mm} \forall j \in \{1,2,..,n\}$ \Comment{Residual error}
\State \hspace{2mm} $S$ = $\phi$ \Comment{Set of selected filters}
\While{$|S| \leq (1-\beta)*n$}
    \For{$i$ in $S^c$}
        \For{$j$ in $\{1,2,..,n\}$}
            \State Compute $Proj_{ij} = R_j.f_{:,i}$
        \EndFor
        \State Total projection $\xi_i \hspace{-1mm}=\hspace{-1mm}\sum_{j=1}^{n}|Proj_{ij}|$
    \EndFor
    \State $ind = \displaystyle\max_i \xi_i$
    \State $S \longleftarrow S \cup \{ind\}$
    \For{$j$ in $\{1,2,..,n\}$}
        \State $\Vec{\lambda}_{j,:} \hspace{-1mm}=\hspace{-1mm} argmin_{\lambda_j} ||f_{:,j} \hspace{-1mm}- \hspace{-1mm}\sum_{l \in S} \hspace{-1mm}\lambda_{j,l} f_{:,l}||^2$
        \State $R_j \longleftarrow f_{:,j} - \sum_{l \in S}({\Vec{\lambda}_{j,:}f_{:,l} })$
    \EndFor
 \EndWhile
 \State \textbf{Output:}
 \State \hspace{2mm} $S$, $\lambda_{j,l} \hspace{2mm}  \forall l\in S \hspace{2mm} \forall j\in \{1,2,..,n\}$  
\end{algorithmic}
\end{algorithm} 
\end{minipage}
\hfill
\begin{minipage}{0.46\textwidth}
\begin{algorithm}[H]
 \caption{Layer Selection: \search}
 \label{algo:search}
 % \footnotesize
 \begin{algorithmic}[1]
 % \scriptsize
 \State \textbf{Input:}  $C$: Number of layers, 
 \State \hspace{11mm} $F_c \hspace{2mm} c=1,...,C$: Filters, 
 \State \hspace{11mm}  $\cD$: Training dataset, 
 \State \hspace{12mm}$\alpha$: Number of filters pruned in one go,
 \State \hspace{12mm}$\beta$: Total pruning ratio over all layers 
 \State \textbf{Initialize:} $y^0_c = F_c^0 * y^0_{c-1} \hspace{2mm} c=1,...,C$, $t\leftarrow 0$
\While{Overall pruning ratio $< \beta$}
    \State $e^t_c = 0  \hspace{2mm} \forall c=1,...,C$
    \State $G_c^t$ $\leftarrow$ FP-OMP($n,\frac{\alpha}{n},F_c^t$) \hspace{-1mm} $\forall c =\hspace{-1mm} 1,...,C$ where $n=|F_c^t|$
    \For{$i = 1,...,|D| $}
    \For{$c = 1,...,C$}
        \State Calculate output $y^{t}_c(i) = F^{t}_c * y^{t}_{c-1}(i) $
        \State $e_c^t = e_c^t + \frac{||y_c^0(i) - G_c^t * y^t_{c-1}(i)||_2}{||y^0_c(i)||_2}$
    \EndFor
    \EndFor
    \State $cmin = \mbox{arg}\min_c e_c$
    \State Revised network params $F^{t+1}_{cmin} = G^t_{cmin}$ ; $F^{t+1}_{c} = F^{t}_{c}\ \forall c\neq cmin$ 
    \State $t \leftarrow t+1$
    \State Run 1 epoch of finetuning.
 \EndWhile
 \State \textbf{Output:} Pruned filters $F^t_c \hspace{2mm} \forall c=1,...,C$
\end{algorithmic}
\end{algorithm}
\end{minipage}
\end{figure}

The above method can be generalized by selecting a set of filters, $S$, in one go. Given the set of selected filters $S$, the error for $j^{th}$ output filter, $f_{:,j} \not\in S$ is given by $\epsilon_j$: 
\begin{equation}
    \label{eq:approxi}
    f_{:,j} = \sum_{l \in S}  \lambda_{j,l}f_{:,l} + \epsilon_j, \forall j \not\in S 
\end{equation}
 The problem of estimating $\lambda_{j,l}, l\in S$ can be posed as a sparse approximation problem:
\begin{equation}
    \label{eq:find_lambda_sparse}
    S^*,\lambda^* =  \mbox{argmin}_{|S|\leq (1-\beta)n,\lambda} \ \hspace{0mm} \sum_{j\in \{1,2,..,n\}} \hspace{-5mm}|| f_{:,j} - \sum_{l \in S}\lambda_{j,l} f_{:,l}||^2 \hspace{1mm}
\end{equation}
where $n$ is the initial number of output channels in the current layer, and pruning ratio $\beta$, is the fraction of channels that are pruned.
Algorithm \ref{algo:omp_sparse} describes an \textit{orthogonal matching pursuit} (OMP) based approximation \citep{tropp2007signal,cai2011orthogonal} for estimating the $S,\lambda_{j,l} \hspace{2mm}\forall j=1,...,n;\ l\in S$.
% Algorithm FP-OMP \citep{10.1145/3564121.3564139} describes an \textit{orthogonal matching pursuit} OMP) based approximation algorithm \citep{tropp2007signal,cai2011orthogonal} for estimating the $S,\lambda_{j,l} \forall j=1,...,n;\ l\in S$.
Note that, equation \ref{eq:find_lambda_sparse} denotes a multi-variate regression version of the sparse approximation problem where the predicted variable is a vector $f_{:,j}, j=1,...,n$ with corresponding independent parameters $\lambda_{j,:}$. Since the total error is the sum of squared errors of the individual components, it is easy to see that projections used in standard OMP algorithm can be replaced with the sum of projections for each multivariate regression component (line 13 of Algorithm \ref{algo:omp_sparse}).
%\tb{Write about the summation formula for calculation of error residual.}
This approach has two advantages: (1) this approach is much faster than LRF since the fine-tuning is performed once for each layer, whereas in LRF it is performed after every pruning step (which is equal to the number of pruned filters in a layer), and (2) this approach provides an optimality guarantee for the selected filters in terms of reconstruction error, under conditions on the incoherence matrix of the features \citep{cai2011orthogonal}.
The overall time complexity of algorithm \ref{algo:omp_sparse} is $\mathcal{O}(|S| n^3)$. In a normal application scenario of uniform pruning, the pruning fraction $\beta$ may be quite high ($\sim 98\%$), resulting in the size of the selected set 
$|S|$ being much smaller than 
$n$, this algorithm is fast ($\mathcal{O}(n^3)$).
% Experimental results confirm that algorithm \ref{algo:omp_sparse} performs better than LRF, both in terms of accuracy as well as pruning time.

LRF also uses a $1\times 1$ convolution layer $g_{j,k}, j,k = 1,...,n$ to compensate for the loss of channel outputs. The modified output feature map, $Z_k, k=1,...,n$ is given by $ Z_k = \sum_{j=1}^n Y_j \ast g_{j,k} := \sum_{j=1}^n X \ast f_{:,j}\ast g_{j,k} $, when the output filters $Y_j, j=1,...,n$ are not pruned. However, after pruning, the output feature map from the original convolutional layer becomes $Y'_j = \sum_{l \in S} X \ast f_{:,l} $. \textit{Weight compensation} is a method for modifying weights for the $1\times 1$ convolutional layer, $g'_{l,k}, l\in S, k=1,...,n$ such that the final predicted output $Z'_k = \sum_{l \in S} X \ast f_{:,l}\ast g'_{l,k}$ matches $Z_k$. The following result provides a formula for calculating $g'_{l,k}$.

\begin{theorem}
Given $Z_k$, $Z'_k$, $g_{j,k}$, and $g'_{l,k}$ defined as above, and $\lambda_{j,l}, j=1,...,n;\ l\in S$ estimated using the filter pruning process. Letting
$ g'_{l,k} = g_{l,k} + \sum_{l' \in S^c} \lambda_{l',l} \ast g_{l',k},  \  \forall l \in S,\ k=1,...,n $, ensures that
$ Z_k - Z'_k = \sum_{l' \in S^c} X \ast \epsilon_{l'} \ast g_{l',k} $, where $\epsilon_{l'}$ is the error vector for the estimation of removed filter $l' \in S^c$, and $S^c$ denotes the set of all removed filters.
\end{theorem}
For brevity, the derivation of the result is provided in the appendix. This result provides us with a formula for updating the weights of the $1\times 1$ filters, thus obviating the need to update them using the SGD update.

% \subsection{Hierarchical Backward Greedy Search (\search) for Whole Network Pruning}
\subsection{Hierarchical Backward Greedy Search (\search) }
\label{sec:HBGS}

The algorithm outlined in the previous section selects a fixed fraction $(1-\beta)$ of filters from each layer. 
However, as shown in the results, each layer can have a different fraction of important filters, depending on the architecture. 
Hence, determining the fraction of filters $\beta_c$ to be pruned in layer $c$ is an important problem. 
%\color{red}Not clear from here, also using the term reconstruction error without defining it--->\color{black} 
%However, $\beta_c$ cannot be determined using the reconstruction error of filter weights as a criterion, since it depends on the rank of the matrix of filter weights $[ f^c_{:,1},...,f^c_{:,n_c} ]$ where $n_c$ is the original number of filters in layer $c$. 
Intuitively, $\beta_c$ should not be determined by the filter-weights since comparing them across layers is not meaningful. For example, the weights in a layer may be scaled by a constant factor, compared to those in another layer.
Hence, we use reconstruction error of filter outputs using input training dataset as the criteria.

Let $\cD=\{(u_1,v_1),...,(u_N,v_N)\}$ be the training dataset, and $y_c(i)$ be the output feature map of layer $c$ when the training datapoint $(u_i,v_i)$ is input to the CNN.  
Also, let $U_c(i)$ be the output of the $c^{th}$ layer when the training datapoint $(u_i,v_i)$ is input to the unpruned CNN. 
%\color{red}Not clear, why we are using $F_c$ instead of $G_c$ in reconstruction error, also is $F_c$ not set of remaining filters?--->\color{black}
Moreover, let $F_c = \{ \sum_{l' \in S_c} (f_{:,l'}^c \ast g_{l',k}^c ), \forall k = 1,...,n_c \}$ be the composite convolutional map of the pruned filters and $1\times 1$ convolution for layer $c$ obtained from a filter pruning method (e.g. FP-OMP described in the previous section). The relative reconstruction error $e_c$ for layer $c$ is given by: 
$e_c = \sum_{(u_i,v_i)\in \cD} \frac{|| U_c(i) - F_c * y_{c-1}(i) ||_2}{||U_c(i)||_2}$.
We propose a \textit{hierarchical backward greedy search} (\search) technique in algorithm \ref{algo:search} to both estimate $\beta_c$ for each layer $c$, as well as select the appropriate filters from each layer. 
Given the number of filters $\alpha$ to be pruned in one go,  the algorithm proceeds iteratively by performing two broad steps in each iteration: (1) determine the optimal $\alpha$ filters to be pruned in each layer $c$, and (2) calculate the total relative reconstruction error $e_c$ as described above.
Finally, the model parameters are updated to prune filters from the layer that leads to the lowest relative reconstruction error.
 Algorithm \ref{algo:search} line 9 describes the first step, and lines 10 - 15 describe an efficient implementation of the second step, where errors for all the layers are computed in one forward pass per example.
 The iterations continue till an overall pruning criterion, e.g. parameter pruning ratio or percentage FLOP reduction is reached.
 The parameter $\alpha$ is chosen to speed up the overall execution and can be chosen as $1$ if the running time is not a concern.
The overall time complexity of algorithm \ref{algo:search}, when using algorithm \ref{algo:omp_sparse} as the filter pruning algorithm is: $\mathcal{O}(T C (N + n^4))$, where $T$ is the number of iterations needed to achieve the desired pruning (depends on $\alpha$ and the pruning criteria), $C$ is the number of layers, $N$ is the training dataset size, and $n$ is the number of filters in each layer. While \search\ (Algorithm \ref{algo:search}) can select a variable number of filters from each layer, the sequential search over the layers for the best filter renders this algorithm expensive. In the next section, we develop a faster filter pruning algorithm.

\subsection{Backward Elimination Algorithm for Filter Pruning}
\label{sec:backward}

The time complexity of the \search\ algorithm depends on the training set size $N$ and the average number of filters per layer $n$. In many cases, when the time complexity of the filter pruning step ($\mathcal{O}( T C n^4)$ is substantially larger than the error computation step $\mathcal{O}(T C N)$, the complexity of the filter pruning algorithm becomes substantially larger than that of the fine-tuning algorithm on the training dataset. The main problem is the OMP-based filter pruning algorithm (FP-OMP) adds one filter in each step, which is an efficient strategy if the number of filters to be selected is small, compared to the total number of filters. 
However, in the context of \search\ algorithm, FP-OMP is sequentially called many times (Algorithm \ref{algo:search} line-9) with decreasing number of filters to be selected each time. In this context, a backward elimination \citep{couvreur2000optimality, ament2021optimality} based approach which iteratively removes the feature which causes a minimal increase in approximation error, is intuitively more appropriate. While the original backward elimination algorithm described in \citep{couvreur2000optimality} is $\mathcal{O}(n^4)$, a faster implementation based on block-matrix inversion was described in \citep{reeves1999efficient}, with time complexity of $\mathcal{O}(n^3)$. Here, we derive a similar algorithm for our problem.

\begin{algorithm}
\caption{Filter Pruning-Backward Elimination (\fpback)}
 \label{algo:backward_sparse}
 % \footnotesize
\begin{algorithmic}[1]
% \scriptsize
 % \State Sparse Approx Problem:
 % \State $f_{j,:}^T \approx A \lambda_{:,j}\ \forall j=1,....,n_c$
 \State \textbf{Input:}  $n$: Number of filters,  $\beta$: Pruning fraction, 
 \State \hspace{11mm} $f_{:,j} \in \RR^{K^2m}$  \hspace{0mm} $j=1,....n$: Filters 

% \State{\textbf{Algorithm:}}
\State \textbf{Initialize:} $S = \{1,...,n\}$ \Comment{Set of currently retained filters}
\State  $B_{:,j} = [ f_{:,j} ]_{j=1,...,n}$ \Comment{Matrix of predicted filter weights}
\State $ A = B $ \Comment{Matrix of retained filter weights}
\While{ $|S| > (1 - \beta) * n$ }
    \State $G = [A^TA]^{-1}\ \in \RR^{|S|\times |S|}$
    \For{$k=1,..., |S|$}    
    \State  $g_k = G_{\{-k\},k}$
    \State $\gamma_k = G[k,k]$ 
    \State $d_k = A_{-k} g_k + a_{k}\gamma_k $ 
    \State $u_k = \frac{\sum_{j=1,...,n} |d_k^T B_{:,j}|^2}{\gamma_k}$
    \EndFor
    \State $k^* = argmin_{k=1,...,|S|} u_k$
    \State $S \longleftarrow  S \setminus \{S[k^*]\}$ \Comment{remove original index corresponding to $k^*$}
    % \State $V  \leftarrow V\backslash V[k^*]$    
    % \State $ n_{curr} = n_{curr} -1$    
    \State $A=A_{:,\{-k^*\} } $  \Comment{remove selected column}
 \EndWhile
\State Calculate $\lambda $ using equation \ref{eq:find_lambda_LS}
% \State $S = \{1,2,...,n_c\} \setminus S'$\\

 \State \textbf{Output:} Set of selected filters-$S$, $\lambda$
\end{algorithmic}
\end{algorithm}    

For simplicity, we follow the notation in \citep{reeves1999efficient}. For a given layer with $n$ output channels, $m$ input channels, and $K\times K$ filter size, we re-define the input filter matrix as $A\in \RR^{K^2m\times n}$, where each column is a flattened vector of filter weights, $A_{:,j} = f_{:,j}, j=1,...,n$. We also denote the output of the sparse approximation as $B\in \RR^{K^2m\times n}$, which is the same as $A$ in this case.
We are interested in the approximation  $B \approx A \mathbf{\lambda}, \mathbf{\lambda}\in \RR^{n\times n}$, where $\lambda_{:j}$ is the weight vector for the $j^{th}$ output filter. We note that the least square solution for $\lambda_{:,j}$ is decoupled from $\lambda_{:,j'}$ where $j\not=j'$.
Hence, the least squares solution  becomes:
\begin{equation}
\begin{aligned}
    \label{eq:find_lambda_LS}
    \lambda_{:,j}^* 
    &= \mbox{argmin}_{\lambda_{:,j}} \  \sum_{j\in \{1,2,..,n\}} || B_{:,j} - \sum_{l=1,...,n} A_{:,l}\lambda_{l,j}||^2  
    &= (A^T A)^{-1} A^T B_{:,j},\ \hspace{5mm}\forall j=1,...,n
\end{aligned}
\end{equation}
Hence, the total least square error is given by:
\begin{equation}
\begin{aligned}
\label{eq:LS_error}
E(A,B) &= \sum_{j=1,...,n} \| B_{:,j} - A (A^T A)^{-1} A^T B_{:,j} \|^2 
      &=  \sum_{j=1,...,n} ( B_{:,j}^T B_{:,j} - B_{:,j}^T A (A^T A)^{-1} A^T B_{:,j} )
\end{aligned}
\end{equation} 

We are interested in calculating the increase in $E(A,B)$ if 1 column of $A$ and the corresponding row of $\lambda$ are removed from the input. Let
% \begin{equation}
% \label{eq:defAminusk}
$A = [A_{-k}\ a_k] \Pi^T_k$,
% \end{equation}
where $A_{-k}$ is the sub-matrix of $A$ after removing the $k^{th}$ column $a_k$, and $\Pi_k$ is the permutation matrix which permutes the columns of $A$ so that $k^{th}$ column is the last. We also have the following definitions of $G_k$, $g_k$, $\gamma_k$ $D_k$ and $d_k$:
\begin{eqnarray}
 \begin{bmatrix}
 G_k & g_k \\
 g_k^T & \gamma_k 
 \end{bmatrix}  &= & \Pi^T_k (A^T A)^{-1} \Pi_k \ ; \ \ \
 \begin{bmatrix}
 D_k^T \\
 d_k^T
 \end{bmatrix} 
  = \begin{bmatrix}
  G_k A_{-k}^T + g_k a_k^T \\
  g_k^T A_{-k}^T + \gamma_k a_k^T
  \end{bmatrix}
\end{eqnarray}

% \begin{eqnarray}
%  (A^T A)^{-1} &= & \Pi_k \begin{bmatrix}
%  G_k & g_k \\
%  g_k^T & \gamma_k 
%  \end{bmatrix} \Pi^T \\
%  (A^T A)^{-1} A^T & = & \Pi_k \begin{bmatrix}
%  D_k^T \\
%  d_k^T
%  \end{bmatrix} 
%   = \Pi_k \begin{bmatrix}
%   G_k A_{-k}^T + g_k a_k^T \\
%   g_k^T A_{-k}^T + \gamma_k a_k^T
%   \end{bmatrix}
% \end{eqnarray}

We note from equation \ref{eq:LS_error}, that only the second term in $E(A,B)$ is dependent on $A$. Hence, we state the following result which connects the above equations to compute the increase in the least square error for the case of \textit{Multivariate Linear Regression}.

\begin{theorem}  Given the definitions of $A_{-k}, d_k$, and $\gamma_k$ above, the following relation holds: 
$\sum_j B_{:,j}^TA_{-k}(A^T_{-k}A_{-k})^{-1}A^T_{-k}B_{:,j}
= \sum_jB_{:,j}^TA(A^TA)^{-1}A^TB_{:,j}-\sum_j\frac{1}{\gamma_k}|d_k^TB_{:,j}|^2$
hence, $E(A_{-k},B) = E(A,B) + \sum_{j=1,...,n} \frac{1}{\gamma_k} |d_k^TB_{:,j}|^2$.
\end{theorem}

This result is a generalization of the result reported in \citep{reeves1999efficient}. For conciseness, we provide the derivation of this result in the appendix. In light of the above result, \fpback\ (algorithm \ref{algo:backward_sparse}) provides the steps for backward elimination-based filter pruning. Note that line 7 in algorithm \ref{algo:backward_sparse} is the most expensive step in the while loop (lines 6 - 17), which can be estimated from $G$ in the previous time-step using block matrix inversion with the worst-case complexity of $O(n^2)$. 
Also, for most calls to the algorithm, the parameter $\beta$ is very low (typically $\leq 0.05$), leading to far fewer iterations of the while loop (lines 6 - 17), which can be assumed to be constant.
Moreover, this cost goes down with iterations as the size of the $G$ matrix reduces significantly with the iterations, reaching only a small fraction for $n$. 
Hence, assuming a constant number of loop executions (lines 6 - 17), the overall complexity of Algorithm \ref{algo:backward_sparse} is $O(n^2)$, which is two orders of magnitude improvement over using algorithm \ref{algo:omp_sparse}.

% \subsection{Hierarchical Backward Greedy Tree Search (HBGTS) for Output-based Layer Selection}
\subsection{Hierarchical Backward Greedy Tree Search (\tree)}
\label{sec:HBGTS}

A key idea behind the hierarchical backward greedy search (\search) algorithm is to select the layer which results in a minimal relative error when pruned from. However, the prediction error of a layer's output is not always indicative of the ultimate predictive performance of the overall network. On the other hand, computing the error of the final network output involves the re-computation of changes through all the downstream layers, starting with the pruned layer. A naive implementation of this can lead to significant compute overhead since it requires $O(CN)$ forward inferences through the network for each pruning step, where $C$ is the number of layers and $N$ is the number of data points in the training set. 

\begin{algorithm}[t]
 \caption{Layer Selection: \tree}
 \label{algo:tree}
 % \footnotesize
 \begin{algorithmic}[1]
 % \scriptsize
 \State \textbf{Input:} $C$: Number of layers, $F_c \hspace{2mm} c=1,...,C$: Filters, $\cD$: Training dataset,
 \State \hspace{12mm} $\alpha$: Number of filters pruned in one step, $\beta$: Pruning ratio over the network
 
 \State \textbf{Initialize:} $y^0_{c,0} = F_c^0 * y^0_{c-1,0} \hspace{2mm} \forall c=1,...,C$, $t\leftarrow 0$

 % \State Calculate $g^0_c$ which are the first set of remaining weights for each layer $c$ using the FP algorithm (e.g. FP-OMP \ref{algo:omp_sparse}) 
 % \State{\textbf{Algorithm:}}
\While{Overall pruning ratio $< \beta$}
% \While{Filtering criteria not reached}
    \State $e^t_{j} = 0 \hspace{3mm} \forall j=0,1,...,C$ \Comment{Initialize total error for each layer pruned}
    \State $G_c^t \leftarrow$ FP-OMP($n,\frac{\alpha}{n},F_c^t$) \hspace{0mm}  $\forall c = 1,...,C$ where $n=|F_c^t|$
    %\State Use FP algorithm on $F_c^t$ (e.g. FP-OMP Algorithm \ref{algo:omp_sparse}) to obtain selected filters $G_c^t,\ \forall c =\hspace{-1mm} 1,...,C$ \\
    \For{$i=1,...,|D|$}
    %\State Update $y^{t}_c(i) = f^{t}_c y^{t}_{c-1}(i) $ for all $c$
    % \State $lst[0] = y_0^t$ \tb{//calculate $0^{th}$ layer output}
    \For{$c$ in $\{1,..,C\}$}
        % \State $temp \gets lst[0]$ 
        \State $y^t_{c,0}(i) = F_c^t * y^t_{c-1,0}(i)$  \Comment{Unpruned propagation of unpruned previous layers} 
        \State $y^t_{c,1}(i) = G_c^t * y^t_{c-1,0}(i)$ \Comment{Pruned propagation of unpruned previous layers} 
        \For{$j=1,...,c-1$}
        \State $y^t_{c,j+1}(i) = F_c^t * y^t_{c-1,j}(i)$ \Comment{Unpruned propagation of previous $j^{th}$  pruned layer} 
        \EndFor
        \EndFor
        \For{$j=1,...,C$} 
        \State $e^t_{j} = e^t_{j} + \frac{||y^t_{C,0}(i) - y^t_{C,C-j+1}(i) ||_2}{|| y^t_{C,0}(i)||_2}$ \Comment{Error in final layer output when $j^{th}$-layer is pruned} 
        \EndFor
    \EndFor
    \State $c_{min} = \mbox{argmin}_{j=1,...,C}(e^t_{j})$
        \State Revised network params: $F^{t+1}_{cmin} = G^t_{cmin}$ ; $F^{t+1}_{c} = F^{t}_{c}\ \forall c\neq cmin$ 
    %\State \textcolor{blue}{New prospective pruned network $g^{t+1}$ becomes}: 
    
    %\State Update $y_c^{t+1}(i) = f^{t+1}_c y^{t+1}_{c-1}(i) $ for all $c \geq cmin$ and $\forall i$
    \State $t \leftarrow t+1$
    \State Run 1 epoch of finetuning for network parameters.
 \EndWhile
 \State \textbf{Output:} Pruned filters $F^t_c,\ \forall c=1,...,C$
    % \State $cmin = |lst| - cmin $
 %    \State Current pruned network becomes $f^{t+1}_{cmin} = g^t_{cmin}$ ; $f^{t+1}_{c} = f^{t}_{c}\ \forall c\neq cmin$ 
 %    \State \textcolor{blue}{New prospective pruned network $g^{t+1}$ becomes}: 
 %    \State Use FP algorithm (e.g., FP-OMP \ref{algo:omp_sparse}) to get the new  $g^{t+1}_{cmin}$; $g^{t+1}_{c} = g^{t}_{c}\ \forall c \neq cmin$
 %    %\State Update $y_c^{t+1}(i) = f^{t+1}_c y^{t+1}_{c-1}(i) $ for all $c \geq cmin$ and $\forall i$
 %    \State $t \leftarrow t+1$
 %    \State Run 1 epoch of finetuning for network parameters.
 % \EndWhile
 %\EndProcedure
\end{algorithmic}
\end{algorithm}

Algorithm \ref{algo:tree} presents \textit{hierarchical backward greedy tree search} (\tree), an efficient implementation for determining the best layer to prune from at every stage. A key idea here is to calculate the error in final layer output $e_j^t$, when layer $j\in \{1,..., C\}$ is pruned, for each input training example. Hence, we need to perform only one forward pass per pruning step for each example $i\in \{1,..., N\}$. To implement this algorithm, we utilize a data structure $y^t_{c,j}$ which stores the output of $c^{th}$ layer $c=1,..., C $, when the $(n-j+1)^{th}$ layer is pruned $j=1,...,C$, in the $t^{th}$ pruning step. Here, $y^t_{c,0}$ represents the output of the $c^{th}$ layer when no filter has been pruned. There are 3 cases while calculating $y^t_{c,:}$ from $y^t_{c-1,:}$ (lines 9, 10, and 12 in algorithm \ref{algo:tree}): (1) calculation of next unpruned output $y^t_{c,0}$ (unpruned propagation of unpruned previous layers), (2) calculation of output corresponding to the current pruned layer $y^t_{c,1}$ (pruned propagation of unpruned previous layers), and (3) unpruned propagation of the pruned outputs corresponding to all the previous layers. Here, we only need to store $y^t_{c,j}$ for the current timestamp $t$. Hence the space complexity of the modified algorithm increases only by $O(nd)$ where $d$ is the output size for each layer. The overall time complexity of the layer selection algorithm for $T$ pruning steps becomes $O(TC^2N)$. Hence, using the backward elimination strategy, the overall time complexity for the proposed algorithm \treeb\ is $O(TC^2N+TCn^2)$. Comparatively, the vanilla backward search version of the algorithm, \searchb\, using backward elimination for filter pruning takes $O(TC^2N+TCn^2)$ time. However, as shown in the next section, \treeb\ marginally outperforms \searchb\ in terms of the accuracy of the pruned models for a given pruning ratio.

\color{black}

%% file: experiment.tex
\begin{table*}[!t]
% \scriptsize
\caption{Performance comparison between different pruning methods on \textbf{VGG16/CIFAR100} at \textbf{98\%} parameter reduction, \textbf{ResNet18/CIFAR10} at \textbf{95\%} parameter reduction, \textbf{ResNet56/CIFAR10} and \textbf{ResNet32/CIFAR10} at \textbf{63\%} parameter reduction, averaged over three runs. ± represents standard deviation, ↓ indicate a drop, and bold/underline denotes the first/second-best result.}
\label{tab:VGG16_ResNet18}
\begin{center}
% \footnotesize
% \scriptsize
\resizebox{\textwidth}{!}{%
\begin{tabular}{l  c c c c|c c c c}

\toprule
&\multicolumn{4}{c|}{\textbf{VGG16/CIFAR10 @ 98\%}}
&\multicolumn{4}{c}{\textbf{ResNet18/CIFAR10 @ 95\%}}\\
\textbf{Method}   

&\multicolumn{1}{c}{\begin{tabular}[c]{@{}c@{}}\textbf{Test Acc}\\ \textbf{(\%)}\end{tabular}} 

&\multicolumn{1}{c}{\begin{tabular}[c]{@{}c@{}}\textbf{Acc ↓}\\ \textbf{(\%)}\end{tabular}} 

&\multicolumn{1}{c}{\begin{tabular}[c]{@{}c@{}}\textbf{Param ↓}\\ \textbf{(\%)}\end{tabular}} 

&\multicolumn{1}{c|}{\begin{tabular}[c]{@{}c@{}}\textbf{FLOPs ↓}\\ \textbf{(\%)}\end{tabular}} 
% \\\hhline{=====}
&\multicolumn{1}{c}{\begin{tabular}[c]{@{}c@{}}\textbf{Test Acc}\\ \textbf{(\%)}\end{tabular}}

&\multicolumn{1}{c}{\begin{tabular}[c]{@{}c@{}}\textbf{Acc ↓}\\ \textbf{(\%)}\end{tabular}} 

&\multicolumn{1}{c}{\begin{tabular}[c]{@{}c@{}}\textbf{Param ↓}\\ \textbf{(\%)}\end{tabular}} 

&\multicolumn{1}{c}{\begin{tabular}[c]{@{}c@{}}\textbf{FLOPs ↓}\\ \textbf{(\%)}\end{tabular}} 
\\\hline

Dense & $93.2 \pm 0.01$    & $0 \pm 0$    & -  & -  & $94.5 \pm 0.02$       & $0 \pm 0$   & -    & -  \\

Random           & $86.3 \pm 0.18$    & $6.9 \pm 0.18$   &98.0   &89.0     & $86.3 \pm 0.06$   & $8.2 \pm 0.06$   &93.7     &85.0     \\
% \hline
% GateDecorators \citep{you2019gate}     & $56.0 \pm 0.34$    & $11.1 \pm 0.34$   & 97.9  & 87.0 & $87.3 \pm 0.09$    & $7.2 \pm 0.09$   & 95.7  & 73.7\\
% % \hline
% EfficientConvNets \citep{li2016pruning} & $29.5 \pm 0.29$    & $37.6 \pm 0.29$    & 98.0  & 86.0 & $70.5 \pm 0.53$    & $24.0 \pm 0.53$    & 95.9  & 79.7\\
% % \hline
% SNAP  \citep{verdenius2020pruning}             & $49.3 \pm 0.44$    & $17.8 \pm 0.44$  & 98.0  & 89.0   & $87.6 \pm 0.94$    & $6.9 \pm 0.94$   & 93.6  & 72.6   \\
% CroP-S \citep{rachwan2022winning}           & $57.4 \pm 0.38$    & $9.7 \pm 0.38$   & 98.0  & 89.0   & $87.5 \pm 0.36$    & $7.0  \pm 0.36$     & 93.6  & 72.3  \\
% CroPit-S \citep{rachwan2022winning}            & $56.5 \pm 0.32$    & $10.6 \pm 0.32$   & 98.1  & 89.0 & $87.8 \pm 0.33$    & $6.7 \pm 0.33$   & 95.0  & 74.5\\ 
% % \hline
% EarlyBird \citep{rachwan2022winning}          & $60.7 \pm 0.31$    & $6.4 \pm 0.31$   & 98.0  & 89.0  & $84.3 \pm 0.32$    & $10.2 \pm 0.32$   & 95.3  & 65.0    \\
EarlyCroP-S \citep{rachwan2022winning}  & $90.0 \pm 0.47$    & $3.2 \pm 0.47$   &98.0   &91.0      & $91.0 \pm 0.52$    & $3.5 \pm 0.52$  &95.1   &85.8    \\

DLRFC \citep{he2022filter}                 & $90.1 \pm 0.07$          & $3.1 \pm 0.07$        & 97.3  & 76.9    & -        & -       & -  & -    \\

SAP \citep{diao2022pruning}                 & -          & -        & -      & -   & $92.4 \pm 0.03$          & $3.1 \pm 0.03$        & 94.9  & 84.9     \\

PL \citep{chen2022over}                 & $90.2 \pm 0.02$          & $3.0 \pm 0.02$        &97.6          &92.0    & -          & -        & -  & -    \\

LRF \citep{joo2021linearly}                 & $90.3 \pm 0.27$          & $2.9 \pm 0.27$        &97.8          &93.0    & $ 91.5\pm 0.37$          & $3.0 \pm 0.37$        &95.1   &85.8     \\

% FP-OMP \citep{10.1145/3564121.3564139}   & $66.4 \pm 0.13$          & $0.7 \pm 0.13$        & 97.9         & 88.0  & $92.9 \pm 0.17$     &  $1.6 \pm 0.17$       & 95.1         & 65.8          \\ 

\fpback   & $91.5 \pm 0.08$          & $1.7 \pm 0.08$        &97.8          &93.0     & $92.8 \pm 0.15$         & $1.7 \pm 0.15$        & 95.1  & 85.8        \\

\search   & $92.7 \pm 0.21$    & $0.5 \pm 0.21$         &98.2           &94.5     & $ 93.9 \pm 0.24$     & $0.6 \pm 0.24$        & 95.3         & 86.2       \\

\searchb   & $92.6 \pm 0.19$    & $0.6 \pm 0.19$         &98.1        &94.3      & $93.7 \pm 0.22$    & $0.8 \pm 0.22$         & 95.2         & 86.0         \\

\tree   & $\textbf{93.5} \pm 0.25$    & $ \textbf{-0.3} \pm 0.25$     &  \textbf{98.6}         &   \textbf{94.8}    & $\textbf{94.7} \pm 0.28$    & $\textbf{-0.2} \pm 0.28$        &  \textbf{95.6}        & \textbf{86.7}       \\

\treeb  & $\underline{93.4} \pm 0.22$    & $\underline{-0.2} \pm 0.22$     &   \underline{98.5}        &  \underline{94.6}  & $\underline{94.6} \pm 0.24$     & $\underline{-0.1} \pm 0.24$     &  \underline{95.4}        &  \underline{86.5}

\\ \bottomrule    
% \\\hline
\\ \toprule

&\multicolumn{4}{c|}{\textbf{VGG16/CIFAR100 @ 98\%}}
&\multicolumn{4}{c}{\textbf{ResNet18/CIFAR100 @ 95\%}}\\
\textbf{Method}   

&\multicolumn{1}{c}{\begin{tabular}[c]{@{}c@{}}\textbf{Test Acc}\\ \textbf{(\%)}\end{tabular}} 

&\multicolumn{1}{c}{\begin{tabular}[c]{@{}c@{}}\textbf{Acc ↓}\\ \textbf{(\%)}\end{tabular}} 

&\multicolumn{1}{c}{\begin{tabular}[c]{@{}c@{}}\textbf{Param ↓}\\ \textbf{(\%)}\end{tabular}} 

&\multicolumn{1}{c|}{\begin{tabular}[c]{@{}c@{}}\textbf{FLOPs ↓}\\ \textbf{(\%)}\end{tabular}} 
% \\\hhline{=====}
&\multicolumn{1}{c}{\begin{tabular}[c]{@{}c@{}}\textbf{Test Acc}\\ \textbf{(\%)}\end{tabular}}

&\multicolumn{1}{c}{\begin{tabular}[c]{@{}c@{}}\textbf{Acc ↓}\\ \textbf{(\%)}\end{tabular}} 

&\multicolumn{1}{c}{\begin{tabular}[c]{@{}c@{}}\textbf{Param ↓}\\ \textbf{(\%)}\end{tabular}} 

&\multicolumn{1}{c}{\begin{tabular}[c]{@{}c@{}}\textbf{FLOPs ↓}\\ \textbf{(\%)}\end{tabular}} 

\\\hline

Dense & $67.1 \pm 0.01$    & $0 \pm 0$    & -  & -  & $68.8 \pm 0.02$       & $0 \pm 0$   & -    & -  \\
Random           & $55.5 \pm 0.16$    & $11.6 \pm 0.16$   & 98.0  & 86.0    & $54.9 \pm 0.13$   & $13.9 \pm 0.13$   &93.1     &84.2     \\
% \hline
% GateDecorators \citep{you2019gate}     & $56.0 \pm 0.34$    & $11.1 \pm 0.34$   & 97.9  & 87.0 & $87.3 \pm 0.09$    & $7.2 \pm 0.09$   & 95.7  & 73.7\\
% % \hline
% EfficientConvNets \citep{li2016pruning} & $29.5 \pm 0.29$    & $37.6 \pm 0.29$    & 98.0  & 86.0 & $70.5 \pm 0.53$    & $24.0 \pm 0.53$    & 95.9  & 79.7\\
% % \hline
% SNAP  \citep{verdenius2020pruning}             & $49.3 \pm 0.44$    & $17.8 \pm 0.44$  & 98.0  & 89.0   & $87.6 \pm 0.94$    & $6.9 \pm 0.94$   & 93.6  & 72.6   \\
% CroP-S \citep{rachwan2022winning}           & $57.4 \pm 0.38$    & $9.7 \pm 0.38$   & 98.0  & 89.0   & $87.5 \pm 0.36$    & $7.0  \pm 0.36$     & 93.6  & 72.3  \\
% CroPit-S \citep{rachwan2022winning}            & $56.5 \pm 0.32$    & $10.6 \pm 0.32$   & 98.1  & 89.0 & $87.8 \pm 0.33$    & $6.7 \pm 0.33$   & 95.0  & 74.5\\ 
% % \hline
% EarlyBird \citep{rachwan2022winning}          & $60.7 \pm 0.31$    & $6.4 \pm 0.31$   & 98.0  & 89.0  & $84.3 \pm 0.32$    & $10.2 \pm 0.32$   & 95.3  & 65.0    \\
EarlyCroP-S \citep{rachwan2022winning}        & $62.8 \pm 0.52$    & $4.3 \pm 0.52$  & 97.9  & 88.0  & $64.1  \pm 0.45$    & $4.7 \pm 0.45$   &94.3   &86.5 \\

DLRFC \citep{he2022filter}                 & $63.5 \pm 0.09$          & $3.56 \pm 0.09$        & 97.1         &  53.7  & -        & -       & -  & -    \\

PL \citep{chen2022over}                 & $63.5 \pm 0.03$          & $3.6 \pm 0.03$        & 97.3         &  87.9  & -          & -        & -  & -    \\

LRF \citep{joo2021linearly}                 & $64.0 \pm 0.31$          & $3.1 \pm 0.31$        & 97.9         &  88.0  & $65.5 \pm 0.29$          & $3.3 \pm 0.29$        &94.6   &87.3     \\

% FP-OMP \citep{10.1145/3564121.3564139}   & $66.4 \pm 0.13$          & $0.7 \pm 0.13$        & 97.9         & 88.0  & $92.9 \pm 0.17$     &  $1.6 \pm 0.17$       & 95.1         & 65.8          \\ 

\fpback   & $66.2 \pm 0.11$          & $0.9 \pm 0.11$        & 97.9         & 88.0    & $67.9 \pm 0.14$         & $0.9 \pm 0.14$        &94.6   &87.3         \\

\search   & $67.3 \pm 0.17$    & $-0.2 \pm 0.17$         & 98.3          & 89.6    & $69.1 \pm 0.19$     & $-0.3 \pm 0.19$        &95.2          &88.5        \\

\searchb   & $67.2 \pm 0.15$    & $-0.1 \pm 0.15$         & 98.1       &   89.4   & $68.9 \pm 0.16$    & $-0.1 \pm 0.16$         &95.1          & 88.3         \\

\tree   & $\textbf{67.8} \pm 0.23$    & $ \textbf{-0.7} \pm 0.23$     &  \textbf{98.5}         &   \textbf{89.8}    & $\textbf{69.7} \pm 0.26$    & $\textbf{-0.9} \pm 0.26$        &  \textbf{95.4}        & \textbf{88.6}       \\

\treeb  & $\underline{67.6} \pm 0.21$    & $\underline{-0.5} \pm 0.21$     &   \underline{98.4}        &  \underline{89.7}  & $\underline{69.5} \pm 0.23$     & $\underline{-0.7} \pm 0.23$     &  \underline{95.3}        &  \underline{88.5}

\\ \bottomrule   

% \\\hline
\\ \toprule

&\multicolumn{4}{c|}{\textbf{ResNet56/CIFAR10 @ 63\%}}
&\multicolumn{4}{c}{\textbf{ResNet32/CIFAR10 @ 63\%}}\\
\textbf{Method}   

&\multicolumn{1}{c}{\begin{tabular}[c]{@{}c@{}}\textbf{Test Acc}\\ \textbf{(\%)}\end{tabular}} 

&\multicolumn{1}{c}{\begin{tabular}[c]{@{}c@{}}\textbf{Acc ↓}\\ \textbf{(\%)}\end{tabular}} 

&\multicolumn{1}{c}{\begin{tabular}[c]{@{}c@{}}\textbf{Param ↓}\\ \textbf{(\%)}\end{tabular}} 

&\multicolumn{1}{c|}{\begin{tabular}[c]{@{}c@{}}\textbf{FLOPs ↓}\\ \textbf{(\%)}\end{tabular}} 
% \\\hhline{=====}
&\multicolumn{1}{c}{\begin{tabular}[c]{@{}c@{}}\textbf{Test Acc}\\ \textbf{(\%)}\end{tabular}}

&\multicolumn{1}{c}{\begin{tabular}[c]{@{}c@{}}\textbf{Acc ↓}\\ \textbf{(\%)}\end{tabular}} 

&\multicolumn{1}{c}{\begin{tabular}[c]{@{}c@{}}\textbf{Param ↓}\\ \textbf{(\%)}\end{tabular}} 

&\multicolumn{1}{c}{\begin{tabular}[c]{@{}c@{}}\textbf{FLOPs ↓}\\ \textbf{(\%)}\end{tabular}} 
\\\hline

Dense 
& $93.45 \pm 0.02$
& $0 \pm 0$
&-
&-
& $92.49 \pm 0.01$
& $0 \pm 0$
&-
&-
\\

SFP \citep{he2018soft}

& $92.91 \pm 0.47$
& $0.54 \pm 0.47$
&63.19
&52.60 
& $91.94 \pm 0.12$
& $0.55 \pm 0.12$
&63.02
&41.50 
\\

FPGM \citep{he2019filter}

& $93.14 \pm 0.21$
& $0.31 \pm 0.21$
&63.21
&52.60 
& $91.79 \pm 0.94$
& $0.70 \pm 0.94$
&63.14
&53.20
\\

HRank \citep{lin2020hrank}

& $92.56 \pm 0.05$
& $ 0.89 \pm 0.05$
& 63.04
& 62.43 
& -
& -
& -
& -
\\

LFPC \citep{he2020learning}

& $92.89 \pm 0.17$
& $0.56 \pm 0.17$
&63.25
&52.90 
& $91.98 \pm 0.06$
& $0.51 \pm 0.06$
&63.05
&52.60
\\

CHIP \citep{sui2021chip}

& $92.88 \pm 0.18$
& $0.57 \pm 0.18$
&63.12
&62.08 
& -
& -
& -
& -
\\

ASyminchange \citep{el2022data} 

& $93.27 \pm 0.11$
& $0.18 \pm 0.11$
&63.28
&62.34 
& -
& -
& -
& - 
\\

LRF \citep{joo2021linearly} 

& $93.49 \pm 0.13$
& $-0.04 \pm 0.13$
&63.35
&62.56 
& $92.52 \pm 0.16$
& $-0.03 \pm 0.16$
&63.34
&62.55 
\\

% FP-OMP \citep{10.1145/3564121.3564139}   

% & $93.91 \pm 0.07$
% & $-0.46 \pm 0.07$
% & 63.35
% & 62.56
% & $92.88 \pm 0.09$
% & $-0.39 \pm 0.09$
% & 63.34
% & 62.55
% \\

\fpback

& $93.88 \pm 0.06$
& $-0.43 \pm 0.06$
& 63.35
& 62.56
& $92.85 \pm 0.04$
& $-0.36 \pm 0.04$
& 63.34
& 62.55
\\

\search

& $94.15 \pm 0.09$
& $-0.70 \pm 0.09$
& 63.87
& 64.91
& $93.06 \pm 0.07$
& $-0.57 \pm 0.07$
& 63.65
& 64.88
\\

\searchb  

& $94.12 \pm 0.08$
& $-0.67 \pm 0.08$
& 63.72
& 64.80
& $93.03 \pm 0.05$
& $-0.54 \pm 0.05$
& 63.63
& 64.78
\\

\tree 

& $\textbf{94.38} \pm 0.13$
& $\textbf{-0.93} \pm 0.13$
& \textbf{63.93}
& \textbf{64.95}
& $\textbf{93.28} \pm 0.12$
& $\textbf{-0.79} \pm 0.12$
& \textbf{63.76}
& \textbf{64.92}
\\ 
\treeb  

& $\underline{94.35} \pm 0.11$
& $\underline{-0.90} \pm 0.11$
& \underline{63.89}
& \underline{64.93}
& $\underline{93.26} \pm 0.09$
& $\underline{-0.77} \pm 0.09$
& \underline{63.75}
& \underline{64.90}

\\ \bottomrule    

% \\\hline
\\ \toprule

&\multicolumn{4}{c|}{\textbf{ResNet56/CIFAR100 @ 98\%}}
&\multicolumn{4}{c}{\textbf{ResNet32/CIFAR100 @ 98\%}}\\
\textbf{Method}   

&\multicolumn{1}{c}{\begin{tabular}[c]{@{}c@{}}\textbf{Test Acc}\\ \textbf{(\%)}\end{tabular}} 

&\multicolumn{1}{c}{\begin{tabular}[c]{@{}c@{}}\textbf{Acc ↓}\\ \textbf{(\%)}\end{tabular}} 

&\multicolumn{1}{c}{\begin{tabular}[c]{@{}c@{}}\textbf{Param ↓}\\ \textbf{(\%)}\end{tabular}} 

&\multicolumn{1}{c|}{\begin{tabular}[c]{@{}c@{}}\textbf{FLOPs ↓}\\ \textbf{(\%)}\end{tabular}} 
% \\\hhline{=====}
&\multicolumn{1}{c}{\begin{tabular}[c]{@{}c@{}}\textbf{Test Acc}\\ \textbf{(\%)}\end{tabular}}

&\multicolumn{1}{c}{\begin{tabular}[c]{@{}c@{}}\textbf{Acc ↓}\\ \textbf{(\%)}\end{tabular}} 

&\multicolumn{1}{c}{\begin{tabular}[c]{@{}c@{}}\textbf{Param ↓}\\ \textbf{(\%)}\end{tabular}} 

&\multicolumn{1}{c}{\begin{tabular}[c]{@{}c@{}}\textbf{FLOPs ↓}\\ \textbf{(\%)}\end{tabular}} 
\\\hline

Dense 
& $69.18 \pm 0.01$
& $0 \pm 0$
&-
&-
& $68.48 \pm 0.02$
& $0 \pm 0$
&-
&-
\\

Random 
& $52.64 \pm 0.18$
& $16.54 \pm 0.18$
&97.90
&87.32
& $58.61 \pm 0.16$
& $9.87 \pm 0.16$
&97.45
&86.90
\\

% SFP \citep{he2018soft}

% & $ \pm $
% & $ \pm $
% &
% & 
% & $ \pm $
% & $ \pm $
% &
% & 
% \\

% FPGM \citep{he2019filter}

% & $ \pm $
% & $ \pm $
% &
% & 
% & $ \pm $
% & $ \pm $
% &
% &
% \\

% HRank \citep{lin2020hrank}

% & $ \pm $
% & $  \pm $
% & 
% &  
% & -
% & -
% & -
% & -
% \\

LFPC \citep{he2020learning}

& $62.83 \pm 0.14$
& $6.35 \pm 0.14$
&97.25
&88.18
& $61.78 \pm 0.13$
& $6.70 \pm 0.13$
&97.13
&87.67
\\

DAIS \citep{guan2022dais}

& $61.23 \pm 0.16$
& $7.95 \pm 0.16$
&97.49
&88.36 
& $60.34 \pm 0.15$
& $8.14 \pm 0.15$
&97.25
&88.12
\\

GCNP \citep{jiang2022channel}

& $62.71 \pm 0.15$
& $6.47 \pm 0.15$
&97.37
&88.42 
& -
& -
& -
& -
\\

DLRFC \citep{he2022filter}

& $62.13 \pm 0.19$
& $7.05 \pm 0.19$
&97.51
&88.73 
& -
& -
& -
& -
\\

% CHIP \citep{sui2021chip}

% & $ \pm $
% & $ \pm $
% &
% & 
% & -
% & -
% & -
% & -
% \\

% ASyminchange \citep{el2022data} 

% & $ \pm $
% & $ \pm $
% &
% & 
% & -
% & -
% & -
% & - 
% \\

LRF \citep{joo2021linearly} 

& $63.73 \pm 0.25$
& $5.45 \pm 0.25$
&97.84
&88.98 
& $63.17 \pm 0.21$
& $5.31 \pm 0.21$
&97.43
&88.56 
\\

% FP-OMP \citep{10.1145/3564121.3564139}   

% & $93.91 \pm 0.07$
% & $-0.46 \pm 0.07$
% & 63.35
% & 62.56
% & $92.88 \pm 0.09$
% & $-0.39 \pm 0.09$
% & 63.34
% & 62.55
% \\

\fpback

& $67.66 \pm 0.10$
& $1.52 \pm 0.10$
& 98.12
& 89.24
& $66.92 \pm 0.07$
& $1.56 \pm 0.07$
& 97.87
& 89.13
\\

\search

& $68.99 \pm 0.15$
& $0.19 \pm 0.15$
& 98.34
& 89.41
& $68.25 \pm 0.14$
& $0.23 \pm 0.14$
& 98.28
& 89.35
\\

\searchb  

& $68.82 \pm 0.13$
& $0.36 \pm 0.13$
& 98.27
& 89.32
& $68.13 \pm 0.11$
& $0.35 \pm 0.11$
& 98.23
& 89.31
\\

\tree 

& $\textbf{69.52} \pm 0.24$
& $\textbf{-0.34} \pm 0.24$
& \textbf{98.61}
& \textbf{89.74}
& $\textbf{68.65} \pm 0.19$
& $\textbf{-0.17} \pm 0.19$
& \textbf{98.46}
& \textbf{89.71}
\\ 
\treeb  

& $\underline{69.39} \pm 0.22$
& $\underline{-0.21} \pm 0.22$
& \underline{98.57}
& \underline{89.68}
& $\underline{68.59} \pm 0.16$
& $\underline{-0.11} \pm 0.16$
& \underline{98.38}
& \underline{89.63}

\\ \bottomrule  

\end{tabular}
}
\end{center}
\end{table*}

\section{Experimental Results}
% In this section, we describe the experimental setup and the datasets used in Section \ref{sec:exp_setup}. We compare the performance of the proposed methods with the state-of-the-art baselines in Section \ref{sec:exp_perf}. We also do a detailed analysis of the working of the proposed method in Section \ref{sec:analyse}.

In this section, we describe the experimental setup and the datasets
used. We compare the performance of the proposed pruning methods against state-of-the-art methods. Furthermore,
a comprehensive examination of the working of the proposed Backward Greedy Search methods is conducted.

\subsection{Experimental setting}
\textbf{Dataset Description:}
For the image classification task, we utilize three datasets: CIFAR10, CIFAR100, and Tiny-Imagenet. CIFAR10 consists of $10$ classes, with a training set of $50$k images and a test set of $10$k images, all with a resolution of $32 \times 32$. Each class contains $5$k training images and $1$k test images. Similarly, CIFAR100 comprises $100$ classes, with $500$ training images and $100$ test images per class. Tiny-Imagenet contains 200 classes and includes a total of 0.1 M images. For our experimentation purpose, we resize the original $64 \times 64$ images of Tiny-Imagenet to $224 \times 224$.

\label{sec:exp_perf}
\textbf{Training Details:} 
% We run our experiments for ResNet18, ResNet32, ResNet56 and VGG16 model architectures on CIFAR10, CIFAR100 and Tiny-Imagenet dataset with different percentage of parameter reduction. We use a pre-trained model while pruning and all the other training settings are adapted from LRF \citep{joo2021linearly}. Using the pre-trained model, we add a warmup of 20 epochs before the pruning starts. Unlike the work of Joo et al. \citep{joo2021linearly} where the network is fine-tuned for an epoch, after each filter removal from a layer, we fine-tune the model after pruning the entire $\beta$ fraction of filters from each layer. Towards the end of pruning the entire model, the network is again fine-tuned for 300 epochs. The initial learning rate for fine-tuning is set to $1e^{-2}$ with a decay of $1e^{-4}$. We also use a step scheduler that divides the learning rate by 10 at epoch 150.
Our experiments involve ResNet18, ResNet32, ResNet56, VGG16, and ResNext models, with various percentages of parameter reduction. We prune a pre-trained model and adopt the training settings from LRF \citep{joo2021linearly}. In contrast to LRF, where the model is fine-tuned for a single epoch after each filter removal, we fine-tune the model after pruning the entire $\beta$ fraction of filters from each layer. After the completion of pruning for the entire model, we fine-tune the pruned model for 300 epochs. For fine-tuning, we set the initial learning rate to $1e^{-2}$ with a decay rate of $1e^{-4}$.
We use a step scheduler that reduces the learning rate by a factor of 10 at epoch 150. Baselines were implemented using code provided by the authors and the recommended hyperparameters were used. We also performed hyperparameter search for the number of epochs, pruning ratios, and learning rates and reproduced the best results.

\textbf{Performance Metric:} We report the \textit{test accuracy} for various pruning methods. The \textit{dense} model's test accuracy corresponds to the \textit{pre-trained} model's accuracy.
Additionally, we report an accuracy drop (\textbf{Acc ↓}) from the dense model. We also report the drop in parameters (\textbf{param ↓}) and FLOPs (\textbf{FLOPs ↓}) as metrics to assess the level of pruning and model efficiency. The reduction in parameters refers to the decrease in the number of parameters/weights across all retained filters. FLOPs, on the other hand, refer to the number of operations (convolutions), within the retained filters.

\subsection{Performance Comparison: Accuracy and Efficiency}
We compare our proposed pruning methods with the state-of-the-art methods in Table \ref{tab:VGG16_ResNet18}. 
% Dense model's test accuracy corresponds to the pre-trained model's accuracy. 
We observe that our proposed methods (\search, \tree, \fpback, \searchb, and \treeb) exhibit higher pruned accuracy compared to state-of-the-art methods for a comparable drop in the number of parameters. 
% Hence, our methods demonstrate a lower accuracy drop from the dense model. 
% We also report the drop in parameters and FLOPs of the pruned model compared to the dense model. 
We also observe that our proposed methods consistently report a higher drop in FLOPs compared to state-of-the-art methods. 
% We divided Table \ref{tab:VGG16_ResNet18} into three partitions, the first partition contains the uniform pruning methods whereas the rest two partitions are for proposed non-uniform pruning methods.
% We note that %\search\ outperforms uniform pruning methods and  \tree\ outperforms \search.  
% We will be using \search\ and \tree\ terms if we make use of Algorithm \ref{algo:omp_sparse} (FP-OMP) in Algorithm \ref{algo:search} and \ref{algo:tree} respectively. 
% Similarly, we will be using \search-B and \tree-B terminology if we make use of Algorithm \ref{algo:backward_sparse} (FP-Backward) in Algorithm \ref{algo:search} and \ref{algo:tree} respectively.

\begin{figure}
\centering
\includegraphics[width=0.9\textwidth]{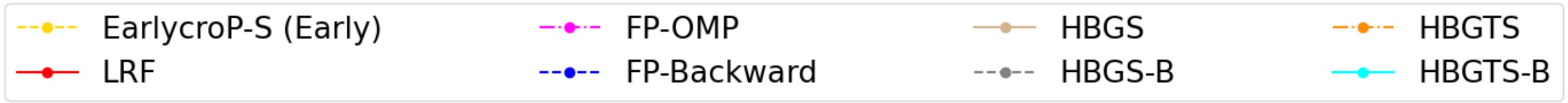}\\
\vspace{1mm}

\subfigure[]{\includegraphics[width=0.33\textwidth]{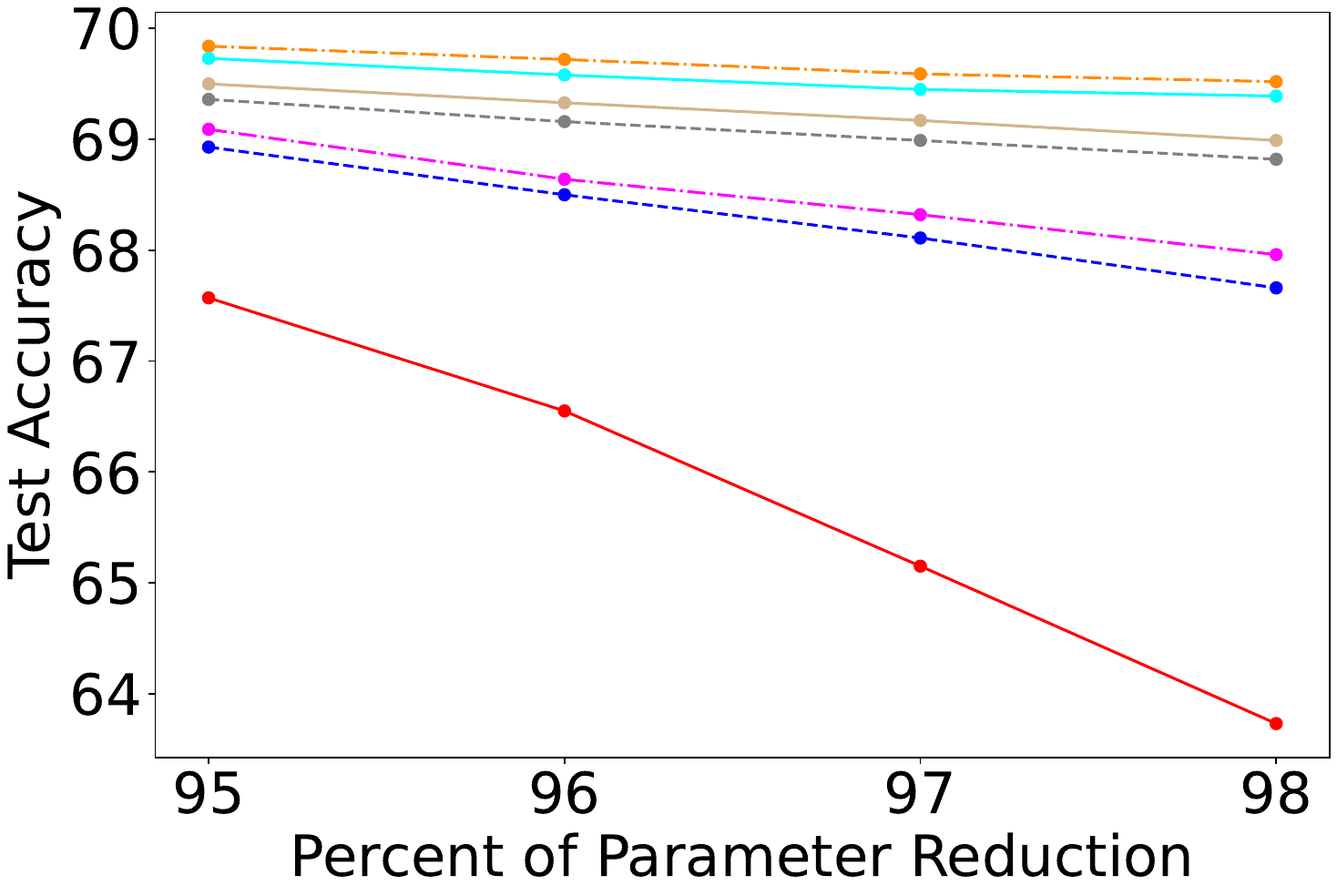}}
\subfigure[]{\includegraphics[width=0.30\textwidth]{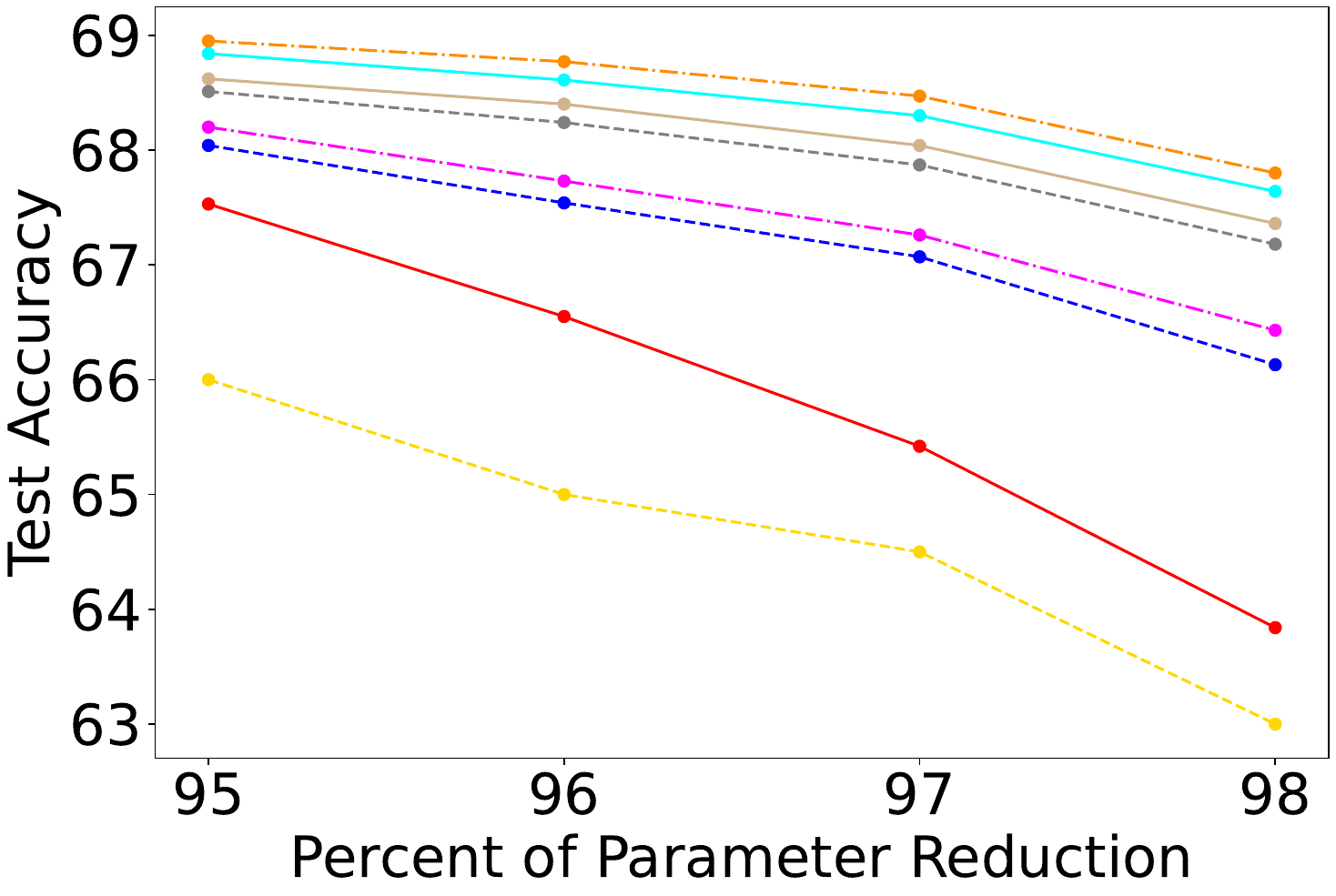}}
\subfigure[]{\includegraphics[width=0.33\textwidth]{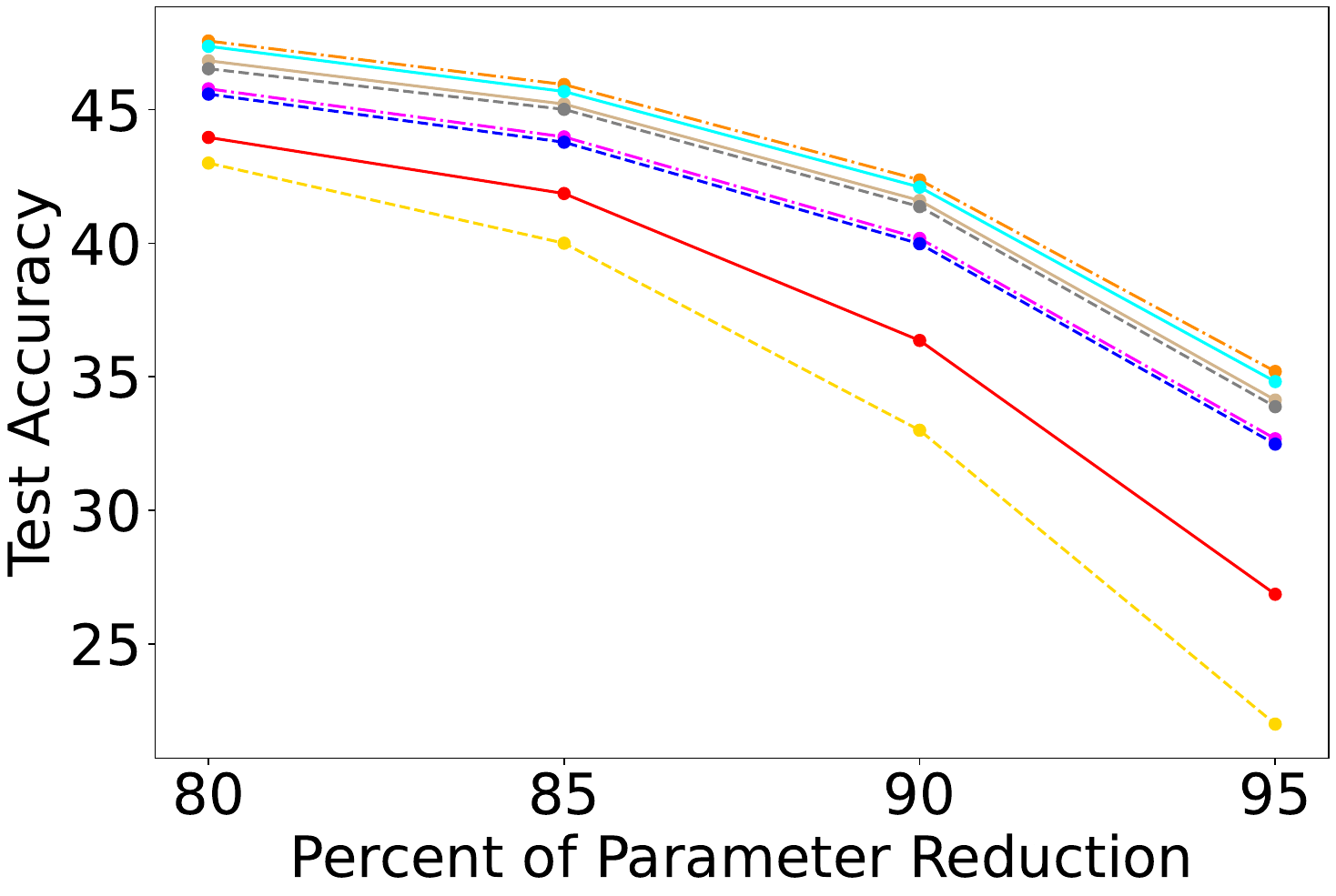}}
%\vspace{-0.3cm}
% \scriptsize
\caption{Test accuracy for (a) \textbf{ResNet56/CIFAR100} (b) \textbf{VGG16/CIFAR100} and (c) \textbf{ResNet18/Tiny-Imagenet} with increasing parameter reduction.}
\label{fig:line_graph}
\end{figure}

\textbf{ResNet and VGG on CIFAR-100 and Tiny-Imagenet:}
Figure \ref{fig:line_graph} and Table \ref{tab:VGG16_ResNet18} provide further insights into the consistently superior performance of our proposed methods.  We observe that the test accuracy of the proposed methods is consistently and significantly better than the test accuracy of the baseline methods. 
The fact that this difference is more pronounced in a difficult dataset (CIFAR100 and Tiny-Imagenet) further demonstrates the superiority of the proposed methods.
From Figure \ref{fig:line_graph}, we notice that, \textit{as the percentage of parameter reduction increases, the difference in test accuracy between our proposed methods and state-of-the-art methods also grows}. \textbf{At higher parameter reduction ($\geq$90\%)}, the proposed methods outperform existing methods by $\sim 3-8\%$ (see Figure \ref{fig:line_graph}). Maintaining or even improving accuracy at such high parameter reduction (>95\%) is very valuable, further highlighting the effectiveness of the proposed methods.
% \textbf{ResNet on Tiny-Imagenet:}
% We observe a similar trend when using ResNet18 for experiments on Tiny-Imagenet (Figure \ref{fig:line_graph}), further confirming the effectiveness of the proposed methods.
% Results shown in Table \ref{tab:VGG16_ResNet18} and Figure \ref{fig:line_graph}, conclude that the difference in test accuracy between our proposed methods and the baseline methods tends to increase with the complexity of datasets, models, and parameter reduction. 
% Further experiments on different models and parameter reduction are shown in the appendix.
% This observation underscores the robustness and effectiveness of our proposed framework.

\textbf{To prune a Large Model: }
Our backward greedy search methods can be used for effectively pruning large models that exceed the capacity of commodity GPUs. We use ResNext101 32x16d as our large model, consisting of 193 M parameters and requires 7.62 GB of GPU memory for loading. Additionally, we use ResNext101 32x8d as our smaller dense model, which has 88 M parameters and requires 3.91 GB for GPU memory.
Table \ref{tab:ResNext_8_16} shows that when ResNext101 32x16d pruned to 98\% parameter reduction using \treeb, achieves a test accuracy that matches its dense counterpart.
Hence, we can efficiently deploy the pruned model on edge devices with GPU memory less than 2GB.
Furthermore, the pruned model takes 5.04 times less GPU memory than the larger dense model. 
Notably, the pruned model even outperforms the smaller dense model, ResNext101 32x8d. 
% Additionally, we introduce ResNext101 32x8d as our smaller dense model, which has 88 M parameters and requires 3.91 GB for GPU memory allocation.

% However, with our methods, we can efficiently deploy this sizable model on edge devices post-pruning. 
\begin{table}[!httb]
\centering
    % \scriptsize
		\caption{Comparison of pruning methods for \textbf{ResNext101 32x16d} (RN16) and a similar sized dense \textbf{ResNext101 32x8d} (RN8) on \textbf{CIFAR10} at \textbf{98\%} parameter reduction.}
		\label{tab:ResNext_8_16}
		% \centering
            % \footnotesize
            % \scriptsize
            % \resizebox{\columnwidth}{!}{%
            \begin{tabular}{l c c c c c}
            \toprule
           
\textbf{Method}   

% & \textbf{Test Acc (\%)}

&\multicolumn{1}{c}{\begin{tabular}[c]{@{}c@{}}\textbf{Test Acc}\\ \textbf{(\%)}\end{tabular}} 

&\multicolumn{1}{c}{\begin{tabular}[c]{@{}c@{}}\textbf{Acc ↓}\\ \textbf{(\%)}\end{tabular}} 

&\multicolumn{1}{c}{\begin{tabular}[c]{@{}c@{}}\textbf{Param ↓}\\ \textbf{(\%)}\end{tabular}} 

&\multicolumn{1}{c}{\begin{tabular}[c]{@{}c@{}}\textbf{FLOPs ↓}\\ \textbf{(\%)}\end{tabular}}  

&\multicolumn{1}{c}{\begin{tabular}[c]{@{}c@{}}\textbf{VRAM}\\ \textbf{(GB)}\end{tabular}}  
            
            \\\hline
            Dense RN16      &  92.1      & 0    & -   & - & 7.62       \\  
            Dense RN8      & 91.8       & 0    & -   &-  & 3.91       \\    
            \fpback         & 92.9      & -0.8    & 98.5        &  89.9  & 1.59        \\ 
            \searchb           & \textbf{93.0}     & -0.9    & 98.7      & 92.1  & 1.55          \\ 
            \treeb             &\textbf{93.2}    & -1.1      & 98.8      & 94.3  & 1.51       \\ 
            \bottomrule          
            \end{tabular}
            % }
\end{table}

\begin{figure}[!t]
    
    \centering
    \subfigure[Uniform Pruning]
    {\includegraphics[width=0.37\textwidth]{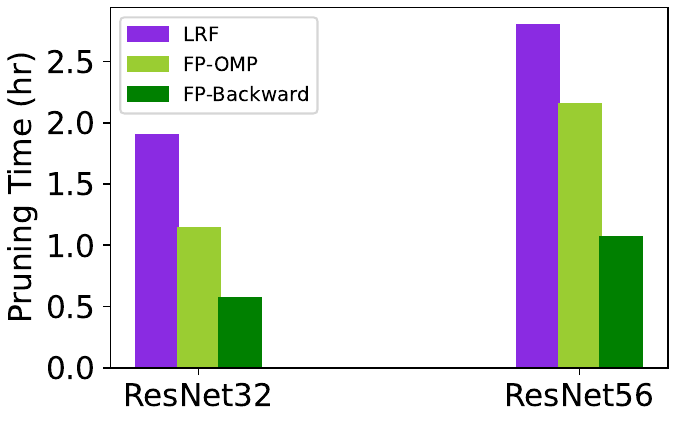}}    
    \hspace{4mm}
    \subfigure[Non-Uniform Pruning]
    {\includegraphics[width=0.37\textwidth]{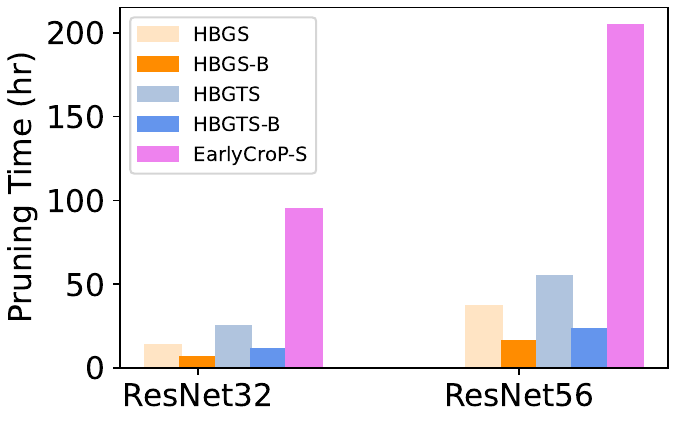}}
    % \scriptsize
  \caption{Time comparison on \textbf{ResNet/CIFAR10} at \textbf{63\%} parameter reduction.}
  \label{fig:bar_time_plot}
  
\end{figure}

\textbf{Time Comparison:} 
Figure \ref{fig:bar_time_plot}(a) provides a comparison of uniform pruning methods in terms of pruning times. We can see that our proposed method \fpback\ is faster than the best baseline FP-OMP by a factor of 2 for a constant pruning ratio in each layer. This is also a fair comparison since the baseline methods also prune a constant fraction of filters from each layer. 

Figure \ref{fig:bar_time_plot}(b) compares the pruning times of various non-uniform pruning methods. Our proposed methods (\search, \searchb, \tree, and \treeb) demonstrate superior computational efficiency compared to the baseline EarlyCroP-S, a non-uniform pruning method. 
Notably, \search\ and \tree\ exhibit higher pruning times relative to \searchb\ and \treeb.
Specifically, \searchb\ achieves a 54.40\% and 56.16\% reduction in pruning time compared to \search\ for ResNet32 and ResNet56, respectively. Likewise, \treeb\ shows a 55.03\% and 57.58\% reduction in pruning time compared to \tree\ for ResNet32 and ResNet56, underscoring the effectiveness of the backward elimination strategy.
Further, as expected \tree\ is computationally more expensive compared to \search\ with approximately double the time. The most efficient hierarchical pruning method (\searchb) takes 5 hours for ResNet32 (see Figure \ref{fig:bar_time_plot}(b)) (for $\alpha=5$, number of filters removed in each round) compared to 1 hour taken by FP-OMP. The increase in time can be further reduced by pruning a higher number of filters ($\alpha$) in each round.

% This is due to the current set of experiments where a batch size of 5 is used for pruning filters. Increasing the batch size would result in a reduction in pruning time. 
% To further decrease the pruning time for \textit{\search} and \textit{\tree}, we propose \textit{\search-B} and \textit{\tree-B}.

% Hence our proposed methods \textit{\search-B} and \textit{\tree-B} are more than two times faster compared to \textit{\search} and \textit{\tree}, respectively. 

%%%%%%%%%%%%%%%%%%%%%%%%%%%%%%%%%%%%%
%%%%%%%%%%%%%%%%%%%%%%%%%%%%%%%%%%%%%
 % \subsection{To prune a Large Model}
%%%%%%%%%%%%%%%%%%%%%%%%%%%%%%%%%%%%%

\subsection{Analysis of Backward Greedy Search Algorithms}
\label{sec:analyse}

\begin{figure}[!b]
    \centering
    \includegraphics[width=0.38\textwidth]{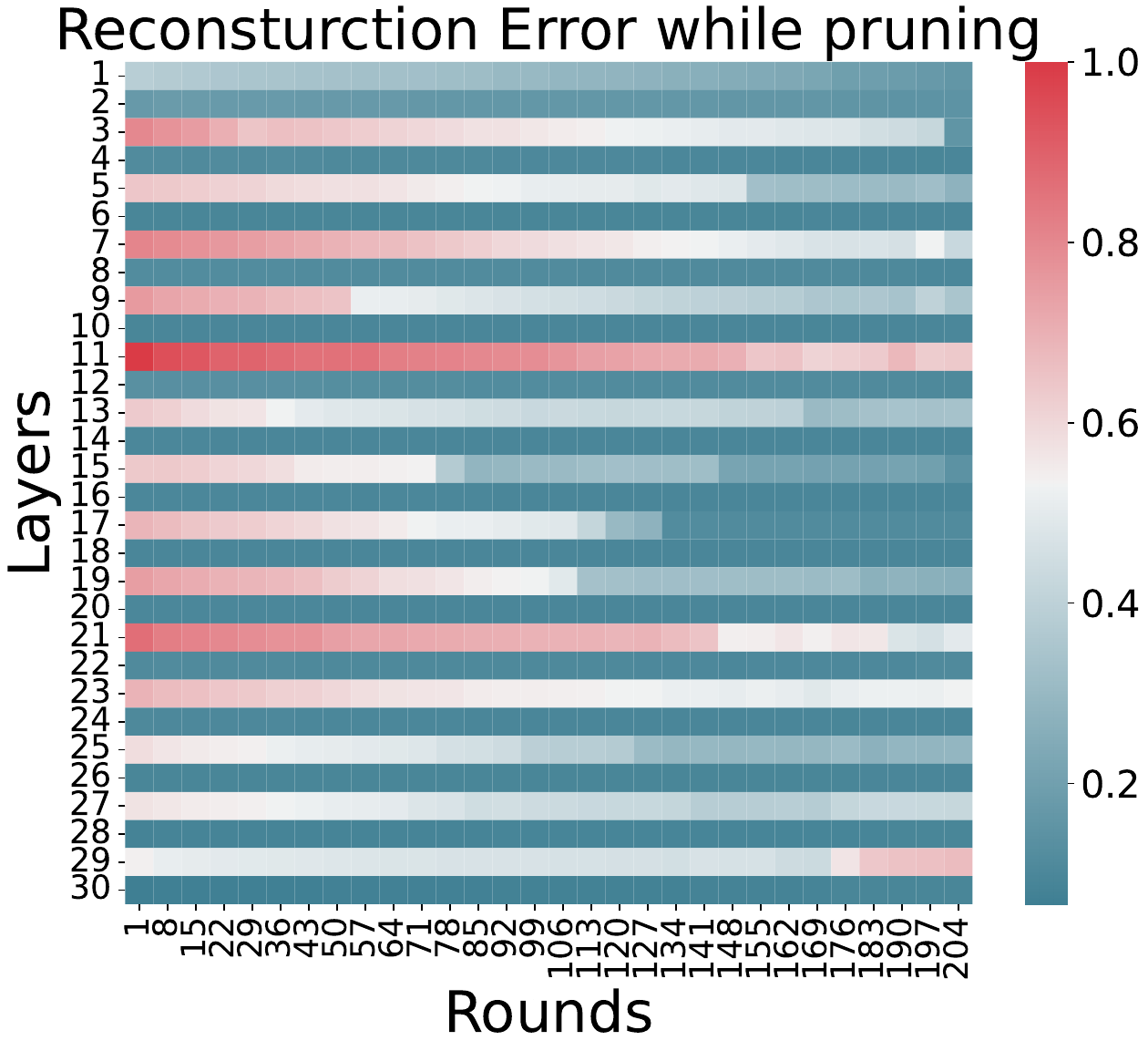}
    \hspace{1cm}
    \includegraphics[width=0.38\textwidth]{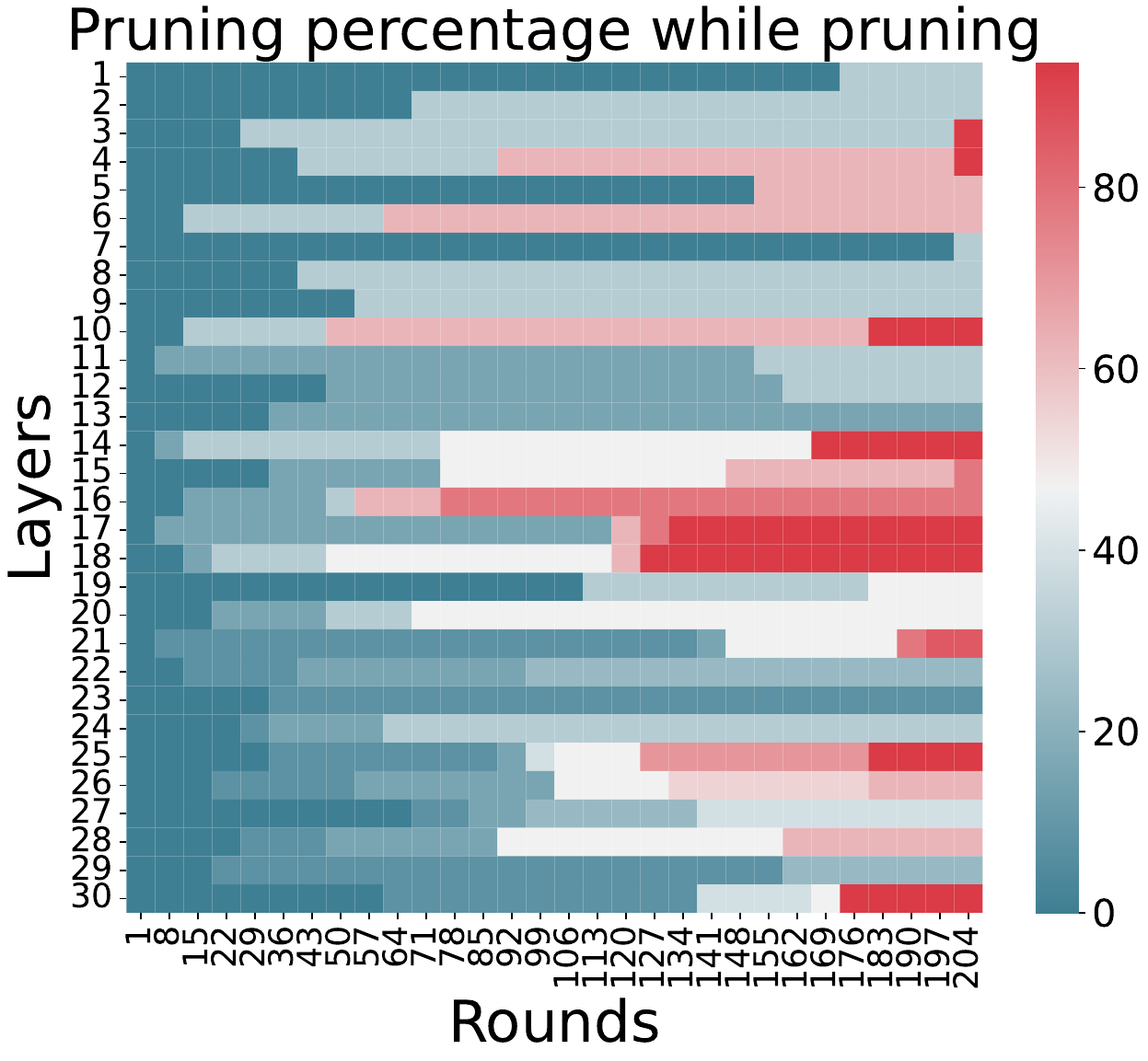}
    % \footnotesize
    % \scriptsize
    % \vspace{-2mm}
    \caption{Heat map for relative reconstruction error and pruning percentage while pruning \textbf{ResNet32} on \textbf{CIFAR100} at \textbf{63\%} parameter reduction.}
    \label{fig:prune_percent_errror_reconst}
\end{figure}

We analyze the working of the proposed greedy search methods in terms of their
pruning quality. 
% To evaluate the pruning quality of the proposed greedy search methods, we analyze their performance. 
Figure \ref{fig:prune_percent_errror_reconst} illustrates a heat map showcasing the relative reconstruction error and the percentage of removed filters for each layer across the pruning rounds, using \treeb\ method for ResNet32 on the CIFAR-100 dataset at 63\% parameter reduction. 
The relative reconstruction error is calculated as $\frac{||y^t_{C,0} - y^t_{C,c} ||_2}{|| y^t_{C,0}||_2}$ where, $y^t_{C,0}$ is the output from the final classification layer when no pruning was done in any of the layers of the network and $y^t_{C,c}$ is the output from the final classification layer when pruning was done at layer c. Both the relative reconstruction error and the pruning percentage are depicted after every $7^{th}$ round, each pruning 5 filters. Examining Figure \ref{fig:prune_percent_errror_reconst}, we observe that the pruning percentage increases with each round, but not uniformly. For example, layers 14 - 18 have higher pruning compared to layers 1-3. Relative reconstruction error also decreases with pruning rounds but is not uniform across layers. From the heat maps, it is evident that our method selects the layer with the least relative reconstruction error for pruning. For example, layers 14 - 18 have moderate relative reconstruction errors in the initial pruning rounds, so the pruning percentage is also not so high for the same. As the pruning rounds increase, the relative reconstruction error decreases for layers 14 - 18 and hence more pruning is done from those layers as visible in the latter rounds in Figure  \ref{fig:prune_percent_errror_reconst}. This is in contrast to uniform pruning approaches, where pruning is uniformly applied across each layer.
%, and the \textit{\tree-B} method prunes a fraction of $\beta$ filters from a specific layer throughout the entire network while retaining the others. 

\begin{figure*}[!t]
\begin{minipage}{0.83\linewidth}
    % \centering
    \includegraphics[width=13.9cm]{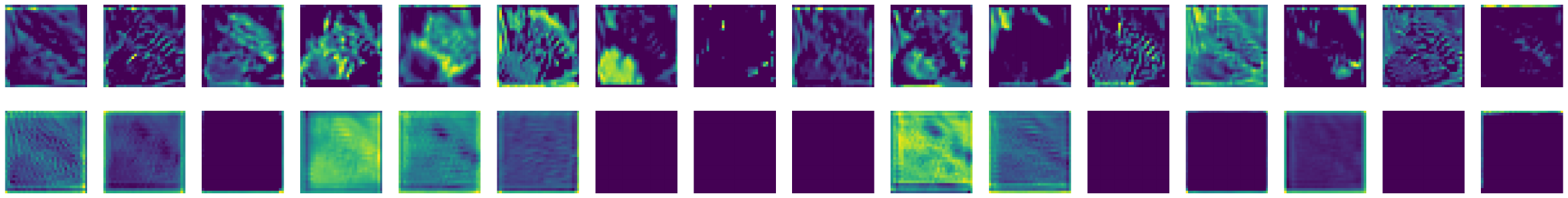}
    % \footnotesize
    % \scriptsize
    % \vspace{-4mm}
    \caption{Visualisation of output feature map of \textbf{ResNet32 $\textbf{2}^{\textbf{nd}}$ layer (top row)} and $\textbf{10}^{\textbf{th}}$ \textbf{layer (bottom row)} on \textbf{CIFAR100}}
    \label{fig:vis}
    \end{minipage}
    \begin{minipage}{0.05\linewidth}
        % \begin{table}
        % \centering
        % \footnotesize
        \vspace{-14mm}
        \begin{small}
        \begin{tabular}{c}
        
            \textbf{31.25\% pruned} \\\\
            \textbf{93.75\% pruned}
        \end{tabular}
        \end{small}
        % \end{table}
    \end{minipage}
\end{figure*}

To understand the intuition behind the filter choices for pruning using \treeb, we present a visualization diagram of feature maps for two layers: Layer 2 (pruned by 31.25\%) and Layer 10 (pruned by 93.75\%). By examining feature maps in Figure \ref{fig:vis} (top row), we can observe that, Layer 2 exhibits a diverse range of filter outputs, indicating their effectiveness in capturing various input features. Consequently, our proposed method prunes only 31.25\% of the filters in Layer 2 (as shown in the last column of pruning percentages in Figure \ref{fig:prune_percent_errror_reconst}). Similarly, Figure \ref{fig:vis} (bottom row) displays feature map outputs from Layer 10, which appear very similar, indicating redundancy in filter outputs. This observation aligns with the pruning percentages shown in the last column of Figure \ref{fig:prune_percent_errror_reconst}, where Layer 10 has 93.75\% of its filters removed. Thus, we can conclude that pruning percentages yielded by \treeb\ are indicative of the amount of information carried by each filter in each layer. Filters with more diverse outputs are retained, while those with redundant outputs are pruned.

%% file: conclude.tex
\section{Conclusion}
% \vspace{-4mm}
In this paper, we propose \search\ and \tree\ as pruning methods that optimize deep neural networks by reducing the number of parameters while maintaining or even improving their accuracy. 
The proposed pruning methods are based on sparse approximation for non-uniform pruning, enabling the removal of redundant filters regardless of the layer they are present in.
We demonstrate our method's consistent performance across various architectures, kernel sizes, and block types through extensive experiments on various datasets. 
We also propose \searchb, and \treeb\ as efficient filter pruning methods, which offer significant advantages in terms of time efficiency compared to \search, and \tree. 
The proposed method can in principle be applied to any linear approximation-based pruning technique, one way of applying it to transformer-based models is pruning the filters in the feed-forward network (FFN). 
% A potential avenue for future research could involve the development of advanced algorithms to determine the optimal number of filters to prune from a layer. Instead of removing filters in arbitrary batch sizes, a more sophisticated approach that dynamically adjusts the pruning rate based on specific layer characteristics or network architecture could potentially yield more optimized results. 

%% file: proofs.tex
\section{Proofs of Results}

\subsection{Weight compensation for multiple channel pruning}
\label{sec:weightcomp}
% \citep{joo2021linearly} had proposed the weight compensation module for a single filter pruning for two purposes: (a) the change in weights of the pruned model will get adjusted by the updation of the $1x1$ convolution and (b) usage of $1x1$ convolution enables the pruning of any network, regardless of its architecture. We adopt this module and derive the compensated weights as per our framework for multiple-channel pruning.
\textbf{Result 1:}
Given $Z_k$, $Z'_k$, $g_{j,k}$, $g'_{l,k}$, and $\lambda_{j,l}, j=1,...,n;\ l\in S$ estimated using the filter pruning process. Letting
$ g'_{l,k} = g_{l,k} + \sum_{l' \in S^c} \lambda_{l',l} \ast g_{l',k},  \  \forall l \in S,\ k=1,...,n $, ensures that
$ Z_k - Z'_k = \sum_{l' \in S^c} X \ast \epsilon_{l'} \ast g_{l',k} $, where $\epsilon_{l'}$ is the error vector for the estimation of removed filter $l' \in S^c$, and $S^c$ denotes the set of all removed filters.

Proof. Consider the input and output of any $K \times K$ convolution layer to be $X = \{X_1 , ... , X_m \}$ and $Y = \{Y_1 ,..., Y_n \}$.
% Since the $1 \times 1$ convolution is built on the top of the $K \times K$ layer, 
$Y$ goes as an input to the $1 \times 1$ convolution. Let the output of the $1 \times 1$ convolution layer be $Z = \{Z_1 ,..., Z_n \}$, followed by $f \in R^{m \times n}$ and $g \in R^{n \times n}$ being the filter weights of $K \times K$ and $1 \times 1$ convolution layer respectively. We can formulate the above setup as:

\begin{equation}            
    \label{eq:kxkoutput}
    Y_j = \sum_{i=1}^m X_i \ast f_{i,j} := X \ast f_{:,j} 
\end{equation} 
\begin{equation}            
    \label{eq:1x1output}
    Z_k = \sum_{j=1}^n Y_j \ast g_{j,k} := \sum_{j=1}^n X \ast f_{:,j}\ast g_{j,k} 
\end{equation} 

Now, let $f_{:,l}$ : $l\in S$ be the selected filter weights and similarly, let $f_{:,l'}$ : $l'\in S'$ be the pruned filter weights. Dividing Equation \ref{eq:1x1output} into the two sets of filter weights, we get:
    
\begin{equation}            
    \label{eq:1x1output_rewrite}
    Z_k = \sum_{l \in S} X \ast f_{:,l}\ast g_{l,k} + \sum_{l' \in S'} X \ast f_{:,l'}\ast g_{l',k} 
\end{equation} 

Following the above terminology, we can write it as:

\begin{equation}            
    \label{eq:lapproxi}
    f_{:,l'} = \sum_{l \in S}  \lambda_{l',l}f_{:,l} + \epsilon_{l'} \hspace{4mm};\forall l' \in S'
\end{equation} 

Substituting Equation \ref{eq:lapproxi} in Equation \ref{eq:1x1output_rewrite}, we rewrite $Z_k$ as $Z'_k$ in terms of retained filter weights $f_{:,l}$:
    
\begin{equation}            
    \label{eq:substituting}
    Z'_k = \sum_{l \in S} X \ast f_{:,l}\ast g_{l,k} + \sum_{l' \in S'} X \ast (\sum_{l \in S}  \lambda_{l',l}f_{:,l} + \epsilon_{l'}) \ast g_{l',k}
\end{equation}

The above can also be re-structured as:

\begin{equation}            
    \label{eq:reordering}
    Z'_k \hspace{-1mm}= \sum_{l \in S} [ X \ast f_{:,l}\ast( g_{l,k} + \sum_{l' \in S'} \lambda_{l',l} \ast g_{l',k})] + \sum_{l' \in S'} X \ast \epsilon_{l'} \ast g_{l',k}
\end{equation}

Once the pruning is performed, Equation \ref{eq:1x1output_rewrite} reduces to 
\begin{equation}
    \sum_{l \in S} X \ast f_{:,l}\ast g_{l,k}
\end{equation}

and Equation \ref{eq:reordering} reduces to 
\begin{equation}
    \label{eq:reduce_eqZk'}
    \sum_{l \in S} [ X \ast f_{:,l}\ast( g_{l,k} + \sum_{l' \in S'} \lambda_{l',l} \ast g_{l',k})]
\end{equation}
Thus, the weight difference after pruning, for $Z_k$ and $Z'_k$, are \newline $\|\sum_{l' \in S'} X \ast f_{:,l'}\ast g_{l',k}\|$ and $\|\sum_{l' \in S'} X \ast \epsilon_{l'} \ast g_{l',k}\|$ respectively. Because $\epsilon_{l'} < f_{:,l'}$, the weight difference in using $Z'_k$ is lesser than that of $Z_k$. Also, the lower the difference in weights, the better the approximation. Hence, we use Equation \ref{eq:reduce_eqZk'} for the weight compensation step to have a lesser weight difference and define the following step:

\begin{equation}            
    \label{eq:new_wc}
    g_{l,k}' = g_{l,k} + \sum_{l' \in S'} \lambda_{l',l} \ast g_{l',k} \hspace{4mm} ;\forall k \in [1,n], \hspace{2mm}\forall l \in S
\end{equation}

For output channel pruning, Equation \ref{eq:new_wc} is re-defined as
\begin{equation}
 g'_{l,:} = g_{l,:} + \sum_{j \in S^c} \lambda_{j,l} \ast g_{j,:}  \hspace{4mm} , \forall l \in S
\end{equation}
while input channel pruning is re-defined as
\begin{equation}
 g'_{:,l} = g_{:,l} + \sum_{j \in S^c} \lambda_{j,l} \ast g_{:,j} \hspace{4mm} , \forall l \in S
\end{equation}

%%%%%%%%%%%%%%%%%%%%%%%%%%%%%%%%%%%%%%%%%%%%%%%%%%%%%%%%%%%%%%%%%%%%%%%%%%%%%%%%%%%%%%%%%%%%%%%%%%%%%%%%%%%%%%%%%%%%%%%%%%%%%%%%%%%%%%%%%%%%%%

\subsection{Backward Elimination Algorithm for Filter Pruning}
\label{sec:multivariate_linear_regress}
\textbf{Result 2:}
Given the definitions of $A_{-k}, d_k$, and $\gamma_k$, the following relation holds: 
$\sum_j B_{:,j}^TA_{-k}(A^T_{-k}A_{-k})^{-1}A^T_{-k}B_{:,j}
= \sum_jB_{:,j}^TA(A^TA)^{-1}A^TB_{:,j}-\sum_j\frac{1}{\gamma_k}|d_k^TB_{:,j}|^2$
hence, $E(A_{-k},B) = E(A,B) + \sum_{j=1,...,n} \frac{1}{\gamma_k} |d_k^TB_{:,j}|^2$.

\textit{Proof.} Given a matrix $A\in \RR^{m\times n}, m \geq n$, with column rank n, and an observation matrix $B \in \RR^{m\times j}$. The best least-squares solution to $A\lambda=B$ with at most r nonzero components is defined as the solution that minimizes the least-squares criterion.
\begin{equation}
\label{eq:least_square}
    Err(\lambda)=\sum_j||B_{:,j}-A\lambda_{:,j}||^2_2
\end{equation}
Unconstrained least-squares solution of $A\lambda=B$ is $\lambda_{:,j}=(A^TA)^{-1}A^TB_{:,j}$. Substituting in Equation \ref{eq:least_square}, we obtain
\begin{equation}
\begin{aligned}
\label{eq:least_square_full}
    Err(\lambda) 
    &=\sum_j ||B_{:,j}-A(A^TA)^{-1}A^TB_{:,j}||^2 \\
    &= \sum_j(B_{:,j}^TB_{:,j}-B_{:,j}^TA(A^TA)^{-1}A^TB_{:,j})
\end{aligned}
\end{equation}
Note that only the second term is a function of A; therefore, maximizing $B_{:,j}^TA(A^TA)^{-1}A^TB_{:,j}$ with respect to combinations of columns comprising A is equivalent to minimizing Equation \ref{eq:least_square} with respect to the combination of nonzero components in the solution.

Let $A_{-k}$ is $A$ with the $k^{th}$ column deleted. $Err_k$ can be written as 
\begin{equation}
    Err_k = \sum_j(B_{:,j}^TB_{:,j}-B_{:,j}^TA_{-k}(A_{-k}^TA_{-k})^{-1}A_{-k}^TB_{:,j})
\end{equation}
A simple update formula for the second term of Equation \ref{eq:least_square_full} can be obtained from \citep{reeves1999efficient}. 
\begin{equation}
\begin{aligned}
    &\sum_jB_{:,j}^TA_{-k}(A_{-k}^TA_{-k})^{-1}A_{-k}^TB_{:,j}\\
    &=\sum_j(B_{:,j}^TA(A^TA)^{-1}A^TB_{:,j}-\frac{1}{\gamma_k}|d_k^TB_{:,j}|^2)
\end{aligned}
\end{equation}
From this result, it is clear that we only need to compare $\sum_j\frac{|d_k^TB_{:,j}|^2}{\gamma_k}$ for all $k$ and eliminate the column whose corresponding value is smallest. Note that $\gamma_k$ is the $k^{th}$ diagonal element of $(A^TA)^{-1}$, and $d_k^TB_{:,j}$ is the $k^{th}$ element of the solution vector $(A^TA)^{-1}A^TB_{:,j}$. Hence we can say,
\begin{equation}
    k^* = min_k(Err_k)
\end{equation}
\begin{equation}
    k^* = min_k \sum_j \frac{|d_k^TB_{:,j}|^2}{\gamma_k}
\end{equation}

%% file: ref.bib
@article{kuzmin2019taxonomy,
  title={Taxonomy and evaluation of structured compression of convolutional neural networks},
  author={Kuzmin, Andrey and Nagel, Markus and Pitre, Saurabh and Pendyam, Sandeep and Blankevoort, Tijmen and Welling, Max},
  journal={arXiv preprint arXiv:1912.09802},
  year={2019}
}

@article{he2018soft,
  title={Soft filter pruning for accelerating deep convolutional neural networks},
  author={He, Yang and Kang, Guoliang and Dong, Xuanyi and Fu, Yanwei and Yang, Yi},
  journal={arXiv preprint arXiv:1808.06866},
  year={2018}
}

@inproceedings{he2020learning,
  title={Learning filter pruning criteria for deep convolutional neural networks acceleration},
  author={He, Yang and Ding, Yuhang and Liu, Ping and Zhu, Linchao and Zhang, Hanwang and Yang, Yi},
  booktitle={Proceedings of the IEEE/CVF conference on computer vision and pattern recognition},
  pages={2009--2018},
  year={2020}
}

@inproceedings{he2019filter,
  title={Filter pruning via geometric median for deep convolutional neural networks acceleration},
  author={He, Yang and Liu, Ping and Wang, Ziwei and Hu, Zhilan and Yang, Yi},
  booktitle={Proceedings of the IEEE/CVF conference on computer vision and pattern recognition},
  pages={4340--4349},
  year={2019}
}

@inproceedings{joo2021linearly,
  title={Linearly replaceable filters for deep network channel pruning},
  author={Joo, Donggyu and Yi, Eojindl and Baek, Sunghyun and Kim, Junmo},
  booktitle={Proceedings of the AAAI Conference on Artificial Intelligence},
  volume={35},
  pages={8021--8029},
  year={2021}
}

@article{zhuang2018discrimination,
  title={Discrimination-aware channel pruning for deep neural networks},
  author={Zhuang, Zhuangwei and Tan, Mingkui and Zhuang, Bohan and Liu, Jing and Guo, Yong and Wu, Qingyao and Huang, Junzhou and Zhu, Jinhui},
  journal={Advances in neural information processing systems},
  volume={31},
  year={2018}
}

@inproceedings{lin2020hrank,
  title={Hrank: Filter pruning using high-rank feature map},
  author={Lin, Mingbao and Ji, Rongrong and Wang, Yan and Zhang, Yichen and Zhang, Baochang and Tian, Yonghong and Shao, Ling},
  booktitle={Proceedings of the IEEE/CVF conference on computer vision and pattern recognition},
  pages={1529--1538},
  year={2020}
}

@article{you2019gate,
  title={Gate decorator: Global filter pruning method for accelerating deep convolutional neural networks},
  author={You, Zhonghui and Yan, Kun and Ye, Jinmian and Ma, Meng and Wang, Ping},
  journal={Advances in neural information processing systems},
  volume={32},
  year={2019}
}

@article{han2015learning,
  title={Learning both weights and connections for efficient neural network},
  author={Han, Song and Pool, Jeff and Tran, John and Dally, William},
  journal={Advances in neural information processing systems},
  volume={28},
  year={2015}
}

@inproceedings{he2017channel,
  title={Channel pruning for accelerating very deep neural networks},
  author={He, Yihui and Zhang, Xiangyu and Sun, Jian},
  booktitle={Proceedings of the IEEE international conference on computer vision},
  pages={1389--1397},
  year={2017}
}

@inproceedings{luo2017thinet,
  title={Thinet: A filter level pruning method for deep neural network compression},
  author={Luo, Jian-Hao and Wu, Jianxin and Lin, Weiyao},
  booktitle={Proceedings of the IEEE international conference on computer vision},
  pages={5058--5066},
  year={2017}
}

@inproceedings{liu2017learning,
  title={Learning efficient convolutional networks through network slimming},
  author={Liu, Zhuang and Li, Jianguo and Shen, Zhiqiang and Huang, Gao and Yan, Shoumeng and Zhang, Changshui},
  booktitle={Proceedings of the IEEE international conference on computer vision},
  pages={2736--2744},
  year={2017}
}

@article{hinton2015distilling,
  title={Distilling the knowledge in a neural network},
  author={Hinton, Geoffrey and Vinyals, Oriol and Dean, Jeff and others},
  journal={arXiv preprint arXiv:1503.02531},
  volume={2},
  number={7},
  year={2015}
}

@inproceedings{tan2019efficientnet,
  title={Efficientnet: Rethinking model scaling for convolutional neural networks},
  author={Tan, Mingxing and Le, Quoc},
  booktitle={International conference on machine learning},
  pages={6105--6114},
  year={2019},
  organization={PMLR}
}

@article{dong2019network,
  title={Network pruning via transformable architecture search},
  author={Dong, Xuanyi and Yang, Yi},
  journal={Advances in Neural Information Processing Systems},
  volume={32},
  year={2019}
}

@inproceedings{fletcher2008robust,
  title={Robust statistics on Riemannian manifolds via the geometric median},
  author={Fletcher, P Thomas and Venkatasubramanian, Suresh and Joshi, Sarang},
  booktitle={2008 IEEE Conference on Computer Vision and Pattern Recognition},
  pages={1--8},
  year={2008},
  organization={IEEE}
}

@article{tropp2007signal,
  title={Signal recovery from random measurements via orthogonal matching pursuit},
  author={Tropp, Joel A and Gilbert, Anna C},
  journal={IEEE Transactions on information theory},
  volume={53},
  number={12},
  pages={4655--4666},
  year={2007},
  publisher={IEEE}
}

@article{cai2011orthogonal,
  title={Orthogonal matching pursuit for sparse signal recovery with noise},
  author={Cai, T Tony and Wang, Lie},
  journal={IEEE Transactions on Information theory},
  volume={57},
  number={7},
  pages={4680--4688},
  year={2011},
  publisher={IEEE}
}

@inproceedings{rachwan2022winning,
  title={Winning the lottery ahead of time: Efficient early network pruning},
  author={Rachwan, John and Z{\"u}gner, Daniel and Charpentier, Bertrand and Geisler, Simon and Ayle, Morgane and G{\"u}nnemann, Stephan},
  booktitle={International Conference on Machine Learning},
  pages={18293--18309},
  year={2022},
  organization={PMLR}
}

@article{frankle2018lottery,
  title={The lottery ticket hypothesis: Finding sparse, trainable neural networks},
  author={Frankle, Jonathan and Carbin, Michael},
  journal={arXiv preprint arXiv:1803.03635},
  year={2018}
}

@inproceedings{10.1145/3564121.3564139,
author = {Purohit, Kiran and Parvathgari, Anurag and Das, Soumi and Bhattacharya, Sourangshu},
title = {Accurate and Efficient Channel Pruning via Orthogonal Matching Pursuit},
year = {2023},
isbn = {9781450398473},
publisher = {Association for Computing Machinery},
address = {New York, NY, USA},
url = {https://doi.org/10.1145/3564121.3564139},
doi = {10.1145/3564121.3564139},
booktitle = {Proceedings of the Second International Conference on AI-ML Systems},
articleno = {16},
numpages = {8},
location = {Bangalore, India},
series = {AIMLSystems '22}
}

@inproceedings{kang2020operation,
  title={Operation-aware soft channel pruning using differentiable masks},
  author={Kang, Minsoo and Han, Bohyung},
  booktitle={International Conference on Machine Learning},
  pages={5122--5131},
  year={2020},
  organization={PMLR}
}

@inproceedings{huang2018data,
  title={Data-driven sparse structure selection for deep neural networks},
  author={Huang, Zehao and Wang, Naiyan},
  booktitle={Proceedings of the European conference on computer vision (ECCV)},
  pages={304--320},
  year={2018}
}

@inproceedings{lin2019towards,
  title={Towards optimal structured cnn pruning via generative adversarial learning},
  author={Lin, Shaohui and Ji, Rongrong and Yan, Chenqian and Zhang, Baochang and Cao, Liujuan and Ye, Qixiang and Huang, Feiyue and Doermann, David},
  booktitle={Proceedings of the IEEE/CVF conference on computer vision and pattern recognition},
  pages={2790--2799},
  year={2019}
}

@article{wen2016learning,
  title={Learning structured sparsity in deep neural networks},
  author={Wen, Wei and Wu, Chunpeng and Wang, Yandan and Chen, Yiran and Li, Hai},
  journal={Advances in neural information processing systems},
  volume={29},
  year={2016}
}

@article{chen2021only,
  title={Only train once: A one-shot neural network training and pruning framework},
  author={Chen, Tianyi and Ji, Bo and Ding, Tianyu and Fang, Biyi and Wang, Guanyi and Zhu, Zhihui and Liang, Luming and Shi, Yixin and Yi, Sheng and Tu, Xiao},
  journal={Advances in Neural Information Processing Systems},
  volume={34},
  pages={19637--19651},
  year={2021}
}

@inproceedings{ament2021optimality,
  title={On the optimality of backward regression: Sparse recovery and subset selection},
  author={Ament, Sebastian and Gomes, Carla},
  booktitle={ICASSP 2021-2021 IEEE International Conference on Acoustics, Speech and Signal Processing (ICASSP)},
  pages={5599--5603},
  year={2021},
  organization={IEEE}
}

@article{couvreur2000optimality,
  title={On the optimality of the backward greedy algorithm for the subset selection problem},
  author={Couvreur, Christophe and Bresler, Yoram},
  journal={SIAM Journal on Matrix Analysis and Applications},
  volume={21},
  number={3},
  pages={797--808},
  year={2000},
  publisher={SIAM}
}

@article{reeves1999efficient,
  title={An efficient implementation of the backward greedy algorithm for sparse signal reconstruction},
  author={Reeves, Stanley J},
  journal={IEEE Signal Processing Letters},
  volume={6},
  number={10},
  pages={266--268},
  year={1999},
  publisher={IEEE}
}

@inproceedings{diao2022pruning,
  title={Pruning Deep Neural Networks from a Sparsity Perspective},
  author={Diao, Enmao and Wang, Ganghua and Zhang, Jiawei and Yang, Yuhong and Ding, Jie and Tarokh, Vahid},
  booktitle={The Eleventh International Conference on Learning Representations},
  year={2023}
}

@article{el2022data,
  title={Data-efficient structured pruning via submodular optimization},
  author={El Halabi, Marwa and Srinivas, Suraj and Lacoste-Julien, Simon},
  journal={Advances in Neural Information Processing Systems},
  volume={35},
  pages={36613--36626},
  year={2022}
}

@article{sui2021chip,
  title={Chip: Channel independence-based pruning for compact neural networks},
  author={Sui, Yang and Yin, Miao and Xie, Yi and Phan, Huy and Aliari Zonouz, Saman and Yuan, Bo},
  journal={Advances in Neural Information Processing Systems},
  volume={34},
  pages={24604--24616},
  year={2021}
}

@inproceedings{murti2022tvsprune,
  title={TVSPrune-Pruning Non-discriminative filters via Total Variation separability of intermediate representations without fine tuning},
  author={Murti, Chaitanya and Narshana, Tanay and Bhattacharyya, Chiranjib},
  booktitle={The Eleventh International Conference on Learning Representations},
  year={2023}
}

@inproceedings{chen2022over,
  title={Over-parameterized Model Optimization with Polyak-$\{$$\backslash$L$\}$ ojasiewicz Condition},
  author={Chen, Yixuan and Shi, Yubin and Dong, Mingzhi and Yang, Xiaochen and Li, Dongsheng and Wang, Yujiang and Dick, Robert P and Lv, Qin and Zhao, Yingying and Yang, Fan and others},
  booktitle={The Eleventh International Conference on Learning Representations},
  year={2023}
}

@inproceedings{he2022filter,
  title={Filter pruning via feature discrimination in deep neural networks},
  author={He, Zhiqiang and Qian, Yaguan and Wang, Yuqi and Wang, Bin and Guan, Xiaohui and Gu, Zhaoquan and Ling, Xiang and Zeng, Shaoning and Wang, Haijiang and Zhou, Wujie},
  booktitle={European Conference on Computer Vision},
  pages={245--261},
  year={2022},
  organization={Springer}
}

@inproceedings{yu2018nisp,
  title={Nisp: Pruning networks using neuron importance score propagation},
  author={Yu, Ruichi and Li, Ang and Chen, Chun-Fu and Lai, Jui-Hsin and Morariu, Vlad I and Han, Xintong and Gao, Mingfei and Lin, Ching-Yung and Davis, Larry S},
  booktitle={Proceedings of the IEEE conference on computer vision and pattern recognition},
  pages={9194--9203},
  year={2018}
}

@inproceedings{goel2020survey,
  title={A survey of methods for low-power deep learning and computer vision},
  author={Goel, Abhinav and Tung, Caleb and Lu, Yung-Hsiang and Thiruvathukal, George K},
  booktitle={2020 IEEE 6th World Forum on Internet of Things (WF-IoT)},
  pages={1--6},
  year={2020},
  organization={IEEE}
}

@article{vadera2022methods,
  title={Methods for pruning deep neural networks},
  author={Vadera, Sunil and Ameen, Salem},
  journal={IEEE Access},
  volume={10},
  pages={63280--63300},
  year={2022},
  publisher={IEEE}
}

@article{he2023structured,
  title={Structured Pruning for Deep Convolutional Neural Networks: A survey},
  author={He, Yang and Xiao, Lingao},
  journal={arXiv preprint arXiv:2303.00566},
  year={2023}
}

@inproceedings{baker2016designing,
  title={Designing Neural Network Architectures using Reinforcement Learning},
  author={Baker, Bowen and Gupta, Otkrist and Naik, Nikhil and Raskar, Ramesh},
  booktitle={International Conference on Learning Representations},
  year={2017}
}

@inproceedings{zoph2016neural,
  title={Neural Architecture Search with Reinforcement Learning},
  author={Zoph, Barret and Le, Quoc},
  booktitle={International Conference on Learning Representations},
  year={2017}
}

@article{hoefler2021sparsity,
  title={Sparsity in deep learning: Pruning and growth for efficient inference and training in neural networks},
  author={Hoefler, Torsten and Alistarh, Dan and Ben-Nun, Tal and Dryden, Nikoli and Peste, Alexandra},
  journal={The Journal of Machine Learning Research},
  volume={22},
  number={1},
  pages={10882--11005},
  year={2021},
  publisher={JMLRORG}
}

@article{lebedev2018speeding,
  title={Speeding-up convolutional neural networks: A survey},
  author={Lebedev, Vadim and Lempitsky, Victor},
  journal={Bulletin of the Polish Academy of Sciences. Technical Sciences},
  volume={66},
  number={6},
  pages={799--811},
  year={2018},
  publisher={Polska Akademia Nauk. Czytelnia Czasopism PAN}
}

@inproceedings{wang2019eigendamage,
  title={Eigendamage: Structured pruning in the kronecker-factored eigenbasis},
  author={Wang, Chaoqi and Grosse, Roger and Fidler, Sanja and Zhang, Guodong},
  booktitle={International conference on machine learning},
  pages={6566--6575},
  year={2019},
  organization={PMLR}
}

@inproceedings{nonnenmacher2021sosp,
  title={SOSP: Efficiently Capturing Global Correlations by Second-Order Structured Pruning},
  author={Nonnenmacher, Manuel and Pfeil, Thomas and Steinwart, Ingo and Reeb, David},
  booktitle={International Conference on Learning Representations},
  year={2021}
}

@inproceedings{molchanov2019importance,
  title={Importance estimation for neural network pruning},
  author={Molchanov, Pavlo and Mallya, Arun and Tyree, Stephen and Frosio, Iuri and Kautz, Jan},
  booktitle={Proceedings of the IEEE/CVF conference on computer vision and pattern recognition},
  pages={11264--11272},
  year={2019}
}

@inproceedings{peng2019collaborative,
  title={Collaborative channel pruning for deep networks},
  author={Peng, Hanyu and Wu, Jiaxiang and Chen, Shifeng and Huang, Junzhou},
  booktitle={International Conference on Machine Learning},
  pages={5113--5122},
  year={2019},
  organization={PMLR}
}

@inproceedings{liu2021group,
  title={Group fisher pruning for practical network compression},
  author={Liu, Liyang and Zhang, Shilong and Kuang, Zhanghui and Zhou, Aojun and Xue, Jing-Hao and Wang, Xinjiang and Chen, Yimin and Yang, Wenming and Liao, Qingmin and Zhang, Wayne},
  booktitle={International Conference on Machine Learning},
  pages={7021--7032},
  year={2021},
  organization={PMLR}
}

@inproceedings{han2015deep,
  title={Deep compression: Compressing deep neural networks with pruning, trained quantization and huffman coding},
  author={Han, Song and Mao, Huizi and Dally, William J},
  booktitle={International Conference on Learning Representations},
  year={2016}
}

@inproceedings{redmon2016you,
  title={You only look once: Unified, real-time object detection},
  author={Redmon, Joseph and Divvala, Santosh and Girshick, Ross and Farhadi, Ali},
  booktitle={Proceedings of the IEEE conference on computer vision and pattern recognition},
  pages={779--788},
  year={2016}
}

@article{minaee2021image,
  title={Image segmentation using deep learning: A survey},
  author={Minaee, Shervin and Boykov, Yuri and Porikli, Fatih and Plaza, Antonio and Kehtarnavaz, Nasser and Terzopoulos, Demetri},
  journal={IEEE transactions on pattern analysis and machine intelligence},
  volume={44},
  number={7},
  pages={3523--3542},
  year={2021},
  publisher={IEEE}
}

@inproceedings{jiang2022channel,
  title={On the Channel Pruning using Graph Convolution Network for Convolutional Neural Network Acceleration.},
  author={Jiang, Di and Cao, Yuan and Yang, Qiang},
  booktitle={IJCAI},
  pages={3107--3113},
  year={2022}
}

@article{guan2022dais,
  title={Dais: Automatic channel pruning via differentiable annealing indicator search},
  author={Guan, Yushuo and Liu, Ning and Zhao, Pengyu and Che, Zhengping and Bian, Kaigui and Wang, Yanzhi and Tang, Jian},
  journal={IEEE Transactions on Neural Networks and Learning Systems},
  year={2022},
  publisher={IEEE}
}

@article{tukan2022pruning,
  title={Pruning neural networks via coresets and convex geometry: Towards no assumptions},
  author={Tukan, Murad and Mualem, Loay and Maalouf, Alaa},
  journal={Advances in Neural Information Processing Systems},
  volume={35},
  pages={38003--38019},
  year={2022}
}

@inproceedings{liebenweinprovable,
  title={Provable Filter Pruning for Efficient Neural Networks},
  author={Liebenwein, Lucas and Baykal, Cenk and Lang, Harry and Feldman, Dan and Rus, Daniela},
  booktitle={International Conference on Learning Representations},
  year={2019}
}

@article{mussay2021data,
  title={Data-independent structured pruning of neural networks via coresets},
  author={Mussay, Ben and Feldman, Dan and Zhou, Samson and Braverman, Vladimir and Osadchy, Margarita},
  journal={IEEE Transactions on Neural Networks and Learning Systems},
  volume={33},
  number={12},
  pages={7829--7841},
  year={2021},
  publisher={IEEE}
}
